\pdfoutput=1
\documentclass[11pt]{article}

\usepackage{rolling_review_files/acl}

\usepackage{times}
\usepackage{latexsym}

\usepackage[T1]{fontenc}
\usepackage[utf8]{inputenc}

\usepackage{microtype}

\usepackage{booktabs}
\usepackage{makecell}
\usepackage{longtable}
\usepackage{graphicx}
\usepackage{multirow}
\usepackage{subcaption}
\usepackage{svg}

\usepackage{amssymb}

\usepackage{algorithm} 
\usepackage{algpseudocode}

\usepackage{pifont}
\title{Text-Guided Image Clustering}

\author{Andreas Stephan$^{1,2,5}$, Lukas Miklautz$^{1,2}$, Kevin Sidak$^{1,2}$, Jan Philip Wahle$^4$, \\
\bf Bela Gipp$^4$,  Claudia Plant$^1$ \and Benjamin Roth$^{1,3}$ \\
$^1$ Faculty of Computer Science, University of Vienna, Austria \\ 
$^2$ UniVie Doctoral School Computer Science, University of Vienna, Austria \\
$^3$ Faculty of Philological and Cultural Studies, University of Vienna, Austria \\
$^4$ Georg-August-Universität Göttingen, Germany \\
\texttt{$^5${andreas.stephan}@univie.ac.at}}

\begin{document}
\maketitle

\begin{abstract}
Image clustering divides a collection of images into meaningful groups, typically interpreted post-hoc via human-given annotations. 
Those are usually in the form of text, begging the question of using text as an abstraction for image clustering. 
Current image clustering methods, however, neglect the use of generated textual descriptions. 
We, therefore, propose \textit{Text-Guided Image Clustering}, i.e., generating text using image captioning and visual question-answering (VQA) models and subsequently clustering the generated text. 
Further, we introduce a novel approach to inject task- or domain knowledge for clustering by prompting VQA models. 
Across eight diverse image clustering datasets, our results show that the obtained text representations often outperform image features. 
Additionally, we propose a counting-based cluster explainability method. 
Our evaluations show that the derived keyword-based explanations describe clusters better than the respective cluster accuracy suggests. 
Overall, this research challenges traditional approaches and paves the way for a paradigm shift in image clustering, using generated text\footnote{Github repo: \url{https://github.com/AndSt/text_guided_cl}}.
\end{abstract}

\section{Introduction}

% Background
Psychologists, neuroscientists, and linguists have long studied the dependence of vision and language in humans \cite{pinker_bloom_1990,nowak2002computational, corballis2017language}.
Although the relationship between these modalities is not fully understood, there is a consistent finding: the brain generates a condensed representation to transmit visual information between brain regions \cite{Cavanagh2021}.
A widely discussed type of representation is often referred to as ``visual language'' or ``language of thought'' \cite{fodor1975language, jackendoff1996language}.
%Arguably, the most universal format to share information between humans is language.
%Related studies confirm that language is a crucial driver of visual understanding. 
Studies based on these concepts suggest that language can be a crucial driver of visual understanding.
For example, children remember conjunctions of visual features better when accompanied by a textual description \cite{Dessalegn2013}, e.g., ``the yellow is left of the black''.
Given this relationship between visual perception and language comprehension, the question arises whether an abstract textual representation benefits image clustering.

\begin{figure}[t]
    \centering
    \includegraphics[width=0.95\columnwidth]{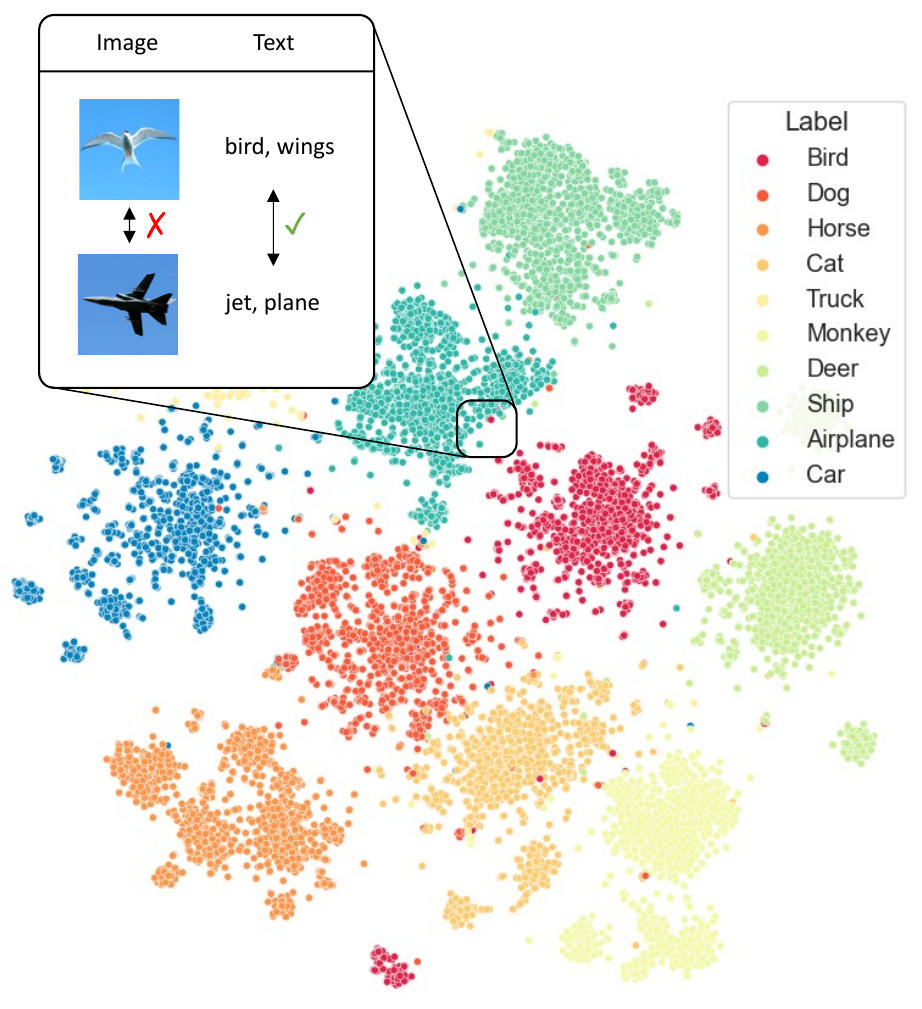}
    \caption{A t-SNE visualization of the BLIP-2 image embeddings for the STL10 dataset. While the images are highly similar (blue background), text such as bird and jet clearly distinguishes objects (and clusters).}
    \label{fig:teaser}
\end{figure}

% Motivate the problem
With the significant growth of visual content created online, image clustering has become essential in, e.g., retrieval systems, image segmentation, or medical applications \citep{img_segmen_survey21,PANDEY2016506,10.1007/978-3-030-88210-5_26}.
%Image clustering is concerned with grouping similar images together and dissimilar images apart \cite{CHIDANANDAGOWDA1978105,elhamifar2013sparse,Chang2017}. 
% Use shared images spaces: CLIP-centroid (Kong2022)
% Others make use of the availability
Language offers dense, human-interpretable information, providing multiple benefits when clustering (Figure \ref{fig:teaser}).
%Here, background information is confused as important.
%Table \ref{tab:pro_con} summarizes properties comparing image and text features for clustering.
%Additionally, language offers a natural interface to guide the clustering process through prompts. 
% Characterize current approaches that tackle the same problem and highlight their shortcomings (as related to your approach)
%Existing multi-modal clustering methods prove the benefits of multimodality \cite{xu2022,cai2022semantic}, but they assume both textual information, e.g. an alt-text, and image are available together.
%Common approaches encode images and texts independently or learn joint feature spaces \cite{loeff2006discriminating,yang2021jecl,wang2020icmsc}. 
%Related to our work, \citet{Kong2022} uses a shared embedding space to find useful keywords.
%Still, they do not derive an abstract textual representation of the image before clustering.
Emerging multi-modal foundation models and large language models (LLMs), e.g., Blip2 \citep{li2023blip2} or GPT-3 \citep{NEURIPS2018_3fd60983}, allow to derive a ``visual language'' from images.

% \input{aux/tables/intro_pro_con_table}

\begin{figure}
    \centering
    \includegraphics[width=0.95\columnwidth]{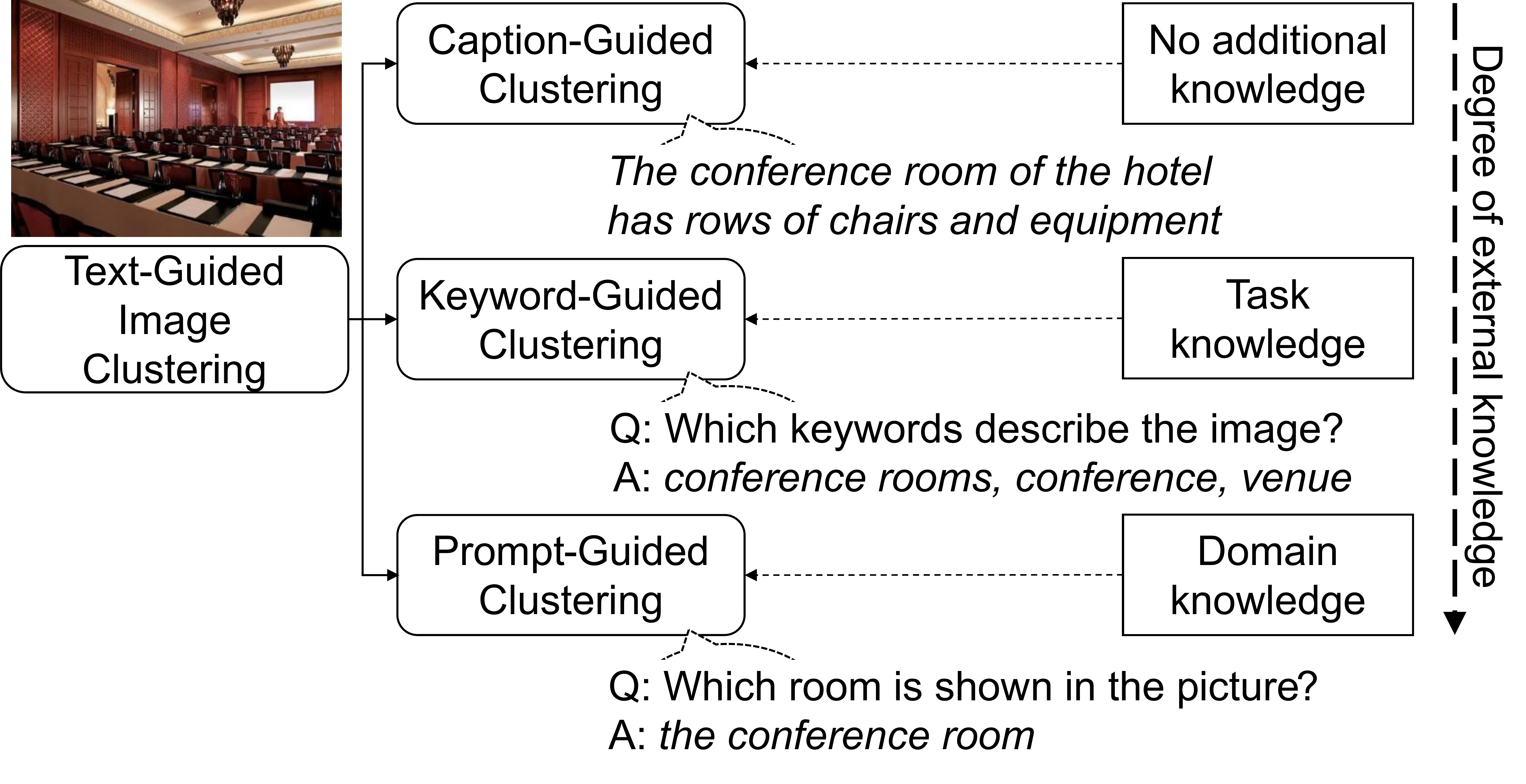}
    \caption{Taxonomy of the text generation processes, structured by the degree of external knowledge. Text is generated BLIP-2 \cite{li2023blip2}.
   % Text is generated from the image (upper left) by BLIP-2 functioning as an image-captioning or VQA model.
    }
    \label{fig:naming}
\end{figure}

% Describe the main novelty of your approach
In this paper, we propose \textit{text-guided image clustering}, i.e., deriving a textual representation from images to perform clustering purely based on their text representation.
% Novelty and Taxonomy
In Figure \ref{fig:naming}, we outline three approaches to text-guided image clustering. 
These approaches are structured by the degree of external knowledge introduced into the clustering process. 

% Caption
First, \textit{caption-guided clustering} uses image captioning models to generate brief descriptions of the image content, requiring no external knowledge.
% Describe your methodology
In order to inspect the qualities of image and text representations, we compare vision encoder embeddings with TF-IDF \citep{sparck1972statistical} and SentenceBERT \citep[SBERT,][]{reimers-2019-sentence-bert} representations of the generated text.
% We choose this setup because image-to-text models typically use large language models to generate text conditioned on the representation of a vision model \citep{alayrac2022flamingo,wang2022git,li2023blip2}. 
Our experiments show that on a broad set of eight image clustering datasets, text representations on average outperform the image representations of three state-of-the-art (SOTA) models.
% Keywords
Second, \textit{keyword-guided clustering} injects knowledge about the clustering task by prompting visual question-answering (VQA) models to generate keywords, using the assumption that only a few keywords of interest are necessary to describe each image sufficiently. 
%Compared to captioned text, results show minor performance improvements for SBERT but a performance increase of on average 5\% for TF-IDF.
Interestingly, we observe an average performance increase of 5\% for TF-IDF-based clusterings.
%This assumption follows from the usage of the chosen data sets in classification tasks, where a single label is used as ground truth.
Third, \textit{prompt-guided clustering} introduces domain knowledge in the form of tailored prompts for VQA models. 
%For example, for images of a scene understanding dataset, applying the prompt ``Which room is shown in the picture?'' generates text providing information about the type of room.
Quantitatively, we observe another performance increase and qualitatively show that clusters related to the question are formed better.
% Explainability
Further, we propose to use the generated text for a straightforward counting-based cluster explainability method, generating a keyword-based description for each cluster.
%Our experiments show quantitatively and qualitatively that the derived descriptions describe clusters better than the respective cluster accuracy suggests.

% Highlight the main insight
Our contributions can be summarized as follows:
\begin{itemize}
\item We propose text-guided image clustering, a novel paradigm leveraging generated text for image clustering. 
%\item We propose text-guided image clustering, a novel paradigm proving that leveraging generated text can outperform clustering solely based on images. We substantiate these findings with a set of comprehensive experiments on a diverse collection of data sets 
% that outperforms image-only representations in our experiments.
%\item We empirically evaluate that a derived visual language for images often is a superior representation %, by comparing the clustering performance of visual encoder representations with the corresponding derived text representations.
% \item By prompting VQA models with task- or domain-specific information, e.g. ``Which activity is shown?'', we inject domain knowledge into the image clustering process. % This can be seen as a new paradigm for the image clustering problem.
\item We introduce a new way of image clustering by injecting task- and domain knowledge via prompting visual question-answering models.
% \item We inject task- and domain knowledge by prompting VQA models, providing a new viewpoint to image clustering. %but realistic viewpoint for the task of image clustering.
%\item Prompt-guided clustering improves the clustering performance on specific subsets of clusters, e.g. rooms vs. all scenes.
% \item Prompt-guided clustering improves the clustering performance for use cases that focus on subsets of clusters, e.g. rooms vs. all scenes, providing the ability to investigate the data using clustering from a specific angle of interest.
%While clustering is often called an ill-defined problem, e.g. time vs. topic axis, our method provides the ability to lessen issues through guidance.
%\item We generate cluster descriptions based on the generated text using a simple, yet strong counting-based cluster explainability method.
\item We show in our experiments that text-guided image clustering is competitive and often outperforms clustering solely based on images on several datasets.
\item We propose a counting-based method to generate a description for each cluster, often exhibiting stronger interpretability than the cluster accuracy suggests.
 % on a wide range of datasets.
%\item We provide an extension to eight existing vision datasets by providing a textual description, i.e. captions, and keywords for each of their examples. 
\end{itemize}

%% Contribs (suggestions Matthias):

% - We propose text-guided image clustering, a novel paradigm proving that leveraging generated text can outperform clustering solely based on images. We substantiate these findings with a set of comprehensive experiments on a diverse collection of data sets (Argument: Text statt Img + SOTA + Experiments)
% - Our method allows injecting task- and domain-specific knowledge via prompting of visual question-answering models into image clustering.
% - This approach strongly enhances the interpretability of the resulting clusters via its counting-based aggregation method to generate descriptions for each cluster.
\section{Related Work}
\label{sec:related_work}

% GOAL: Beschreibung des related works, sodass die Frage aufkommt: "Warum hat noch niemand den generierten Text für Clustering genutzt"?
We approach image clustering in a novel way by generating more abstract text descriptions using image-to-text models. Therefore, we discuss how our approach relates to earlier work in image clustering (Section \ref{subsec:image_clustering}), text clustering (Section \ref{subsec:text_clustering}) and give an overview of the enabling technology of image-to-text models in Section \ref{subsec:image_text_models}.

\subsection{Image Clustering}
\label{subsec:image_clustering}
%TODO: Shortly cite kong, https://openreview.net/pdf?id=-JW-1Fg-v2; Semantic enhanced image clustering

Clustering is the task of grouping similar objects together while keeping dissimilar ones apart. A key problem for unsupervised clustering of images is finding a good similarity measure. Deep learning-based clustering methods approach this problem by learning a representation that maps semantically similar images closer together \cite{dec,dcn,gat_cluster,deep_cluster_caron,deep_clustering_survey}. 
A downside of unsupervised methods is that relying only on image information can suffer from the \textit{blue sky problem} \cite{blue_sky}. 
For example, in Figure \ref{fig:teaser}, the blue background pixels make up most of the images. Our approach circumvents this downside by generating a concise textual description of an image. % that captures the salient features better.
%Otherwise, e.g., when clustering in pixel space directly, an algorithm might mostly focus on the highest amount pixels that make up an image, which is usually the background.
%For example, clustering images of birds and planes can be difficult as both of them share the same background (a blue sky), which makes up most of the image. 
%This transfers to other cases in image clustering, where the background is more prominent than the object of interest. %This example is also known as the \textit{blue sky problem} \cite{x}.
%While image clustering methods rely solely on a representation learned from images suffer from the blue sky problem, 
Multi-view clustering methods like \citep{jin2015cross, chaudhary2019novel,yang2021jecl,xu2022} combine heterogeneous views of data instances into a single clustering. In contrast to our work, they assume the availability of all modalities. %, including possible text descriptions and the correct relation of the given modalities.  
%Multi-view clustering methods like \citep{jin2015cross, chaudhary2019novel,yang2021jecl,xu2022} show that text can be a valuable addition for image clustering, but they consider also views from other modalities. 
%Avoiding the need for text \citet{Kong2022} use CLIP \citep{pmlr-v139-radford21a} to generate text centroids assigned by the image features and perform clustering on the shared encoder representations. 

% Multi-Modal / Multi-View Clustering
% XAI part:
An important problem in clustering is explainability \citep{Fraiman2011InterpretableCU,pmlr-v119-moshkovitz20a}, aiming to describe the content of the individual clusters.
In general, there are clustering algorithms designed such that the resulting clustering is explainable \citep{ijcai2018p176}, or post-processing methods that explain a given clustering.
Existing methods use interpretable features such as semantic tags \citep{Sambaturu_Gupta_Davidson_Ravi_Vullikanti_Warren_2020,NEURIPS2018_3fd60983}, especially when textual explainability is considered.
For instance, \citet{ijcai2021p460} uses integer linear programming to assign tags to clusters.
Contrary to our approach, these methods assume given textual tags.
% TODO: discuss prototype: https://link.springer.com/chapter/10.1007/978-3-031-19775-8_21

\subsection{Text Clustering}
\label{subsec:text_clustering}
% TODO: Maybe ChatGPT-Guidance, ...

Typically, the text is transformed into a vector representation, and then a clustering algorithm, e.g., K-Means, is applied.
Early text representation approaches use counting-based representations such as Bag-of-Words (BoW) or TF-IDF \citep{sparck1972statistical,zhang2011comparative}.
The field moved away from frequency-based approaches as they neglect word order and cannot represent contextualized information, e.g., computer `mouse' vs. the animal `mouse' \citep{peters-etal-2018-deep}.
In recent years, the focus in Natural Language Processing (NLP) shifted towards contextualized neural network-based vector encodings, dominated by transformer-based methods \citep{vaswani_attention_2017}. 
The first breakthrough in transformer-based sentence representations was Sentence-BERT (SBERT) \citep{reimers-2019-sentence-bert}, a siamese network architecture fine-tuning BERT \cite{devlin-etal-2019-bert} on supervised datasets, e.g. NLI.
Following SBERT, text representation techniques are mostly trained using contrastive learning where the choice of positive and negative pairs is unsupervised, e.g., SimCSE \citep{gao-etal-2021-simcse}, or weakly-supervised, e.g., E5 \citep{wang2022text}.

% Text clustering methods group similar documents together while keeping dissimilar ones apart. Topic modelling is a special case of text clustering in which documents are grouped together if they share the same topic.

% Recent breakthroughs in large language models (LLMs) have enabled strong topic models, like BERTopic \cite{bertopic}. BERTopic uses pretrained sentence embedding models (SentenceBERT \cite{x}) to embedd the text and to cluster it. We make use of SentenceBERT as well to embedd the generated image descriptions.

\subsection{Image-To-Text Models}
\label{subsec:image_text_models}

%Image captioning is the task concerned with generating a textual description of an image.

%- Add virst multi-modal captioning models -> ... -> got better -> VQA

%- add git expl.
%- add blip2 expl.: no multimodal finetuning of encoders
%- maybe small discussion about relation
%- datasets for training: scraped / labeled / filtered

% - Paar citations für anfängliche Image-to-text (histories), 2-5; VQA 2015 (Anton et-al)
% - Introduce Models used (Flamingo; GIT, BLIP2)
% - Datasets 
% Flamingo, 1-2 oveerview papers 

Image captioning %, an integral task in image-to-text models, 
provides textual descriptions for given images. 
NIC~\citep{43274} introduces the now common use of an image encoder and a language decoder.
Subsequent models~\citep{Radford2021LearningTV, Yuan2021FlorenceAN} additionally allow multi-modal inputs, integrating both image and textual information to improve captioning and support tasks like Visual Question Answering (VQA)~\citep{Antol_2015_ICCV}. 
%Despite these advancements, one major problem remaining is the number of trainable parameters in large-scale models. 
\citet{wang2022git} use only one image encoder and one text decoder, and perform image /video captioning and VQA in one simplified architecture.
Flamingo~\citep{alayrac2022flamingo} allows interleaving images and text by introducing Perceiver Resamplers on top of pre-trained image and language models.
BLIP-2~\cite{li2023blip2} is a state-of-the-art model that takes fixed pre-trained language and image models and only fine-tunes a so-called Query-Transformer, which only uses a few trainable parameters.
This is useful in our experiments because the underlying models are not trained on multimodal data, ensuring a fair comparison of the respective representations.

\section{Methodology}
\label{sec:methodology}
%\section{Setup}

% For the rest of the paper, we use the following datasets and methods. Find a more comprehensive overview, including dataset statistics in Appendix %TODO

% The goal of this paper is to compare image and generated text representations for image clustering. 
% The following describes the introduced problem setting.

We describe the formal setup, the experimental setup, and the chosen datasets.

\subsection{Problem Definition} 
\label{subsec:problem_def}

Let $\mathbf{X} = {\mathbf{x}_1, \cdots, \mathbf{x}_n} \subset \mathcal{X}$ denote the set of images in our dataset. 
The goal of image clustering is to obtain a clustering $h: \mathcal{X} \rightarrow \mathcal{Y}$ that assigns images to their respective clusters. 
We propose to employ image-to-text models which typically consist of an image encoder $f: \mathcal{X} \rightarrow \mathcal{Z}$, embedding images into a latent space $\mathcal{Z} \subset \mathbb{R}^d$, and a text decoder, i.e. a LLM, $g: \mathcal{Z} \rightarrow \mathcal{T}$, where $\mathcal{T}$ is some text space.
The text is subsequently embedded $t: \mathcal{T} \rightarrow \mathcal{V} \subset \mathbb{R}^l$ and clustered, e.g., with K-Means.

\subsection{Experimental Setup}
\label{subsec:exp_setup}

%The goal of this paper is to compare representations of images and generated text for image clustering.
The central goal of this paper is to compare representations based on images and generated text for the task of image clustering.
The following describes the choices and evaluation criteria common to all experiments. 

\noindent\textbf{Clustering.} 
To shed light on the question of whether text is a (more) suitable representation for image clustering, we compare the performance of a clustering on the image space $\mathbf{Z} = f(\mathbf{X})$ and of a clustering on the vectorization of the generated text $\mathbf{T} = t(g(\mathbf{Z}))$. 
Following the deep clustering  \cite{dec,dcn} and self-supervised learning \cite{iBot} literature, we use K-Means to evaluate the suitability of the respective image and text embeddings for clustering.
We run K-Means 50 times in all experiments and report the mean outcome to get robust results. 
Whenever we need a single run, e.g., for qualitative analysis, the run with the lowest K-Means loss, also called inertia, is used.

\noindent\textbf{Vectorization. } 
In order to employ clustering algorithms, images and texts need to be represented as vectors. 
For image vectorization, we use the latent space of an image encoder. 
We experiment with multiple models introduced in Section \ref{subsec:cgic}.
For text vectorization, one frequency-based and one neural algorithm are considered. 
TF-IDF \cite{sparck1972statistical} is a standard counting-based representation. 
Using the scikit-learn \citep{scikit-learn} implementation, English stop-words are removed, and a maximum vocabulary of 2000 words is set. No additional preprocessing is performed.
Since nowadays transformer-based text representations are the standard, we experiment with SBERT\footnote{\url{https://huggingface.co/sentence-transformers/all-MiniLM-L6-v2}} \citep{reimers-2019-sentence-bert} as it was the first BERT-based sentence representation.
Note that larger, newer, and better transformer-based models are available. We deliberately choose a widely used, competitive, small model as this strengthens our claim that clusterings based on generated text often outperform clusterings based on image representations.

% Note that we do not compare to Multi-View Clustering because there the question is how to use multiple (available) modalities for clustering. While these methods are applicable to our generated text, it is not the focus of this work.

%It is important that we do not compare our approach with multiview clustering techniques, which typically assume the availability of both text and image data and utilizes both modalities. We do not assume the availability of text a priori.
%Instead, our investigation solely concentrates on exploring the potential benefits and qualities of (generated) text.
%Multiview: (I + T) -> C
% We: I -> C vs. T -> C
% => Not consensus because (triangle different)

\noindent\textbf{Metrics.}
To measure clustering performance, the Normalized Mutual Information (NMI) \citep{JMLR:v11:vinh10a} and the Cluster Accuracy (Acc) \cite{cluster_acc} are computed. 
Both metrics take values between 0 and 1, where higher numbers indicate a better match with the ground truth labels. 
For the sake of readability, we multiply them by 100.

\subsection{Datasets}

We consider a diverse collection of datasets, separated into three groups according to various challenges. 
Partially, there is an overlap between the properties of the datasets. Nevertheless, our selection of datasets is motivated by this grouping. Note that this is a more diverse set of datasets as typically used \cite{9878974,Qian_2023_ICCV}.
An overview of the dataset statistics and samples of each dataset are depicted in Tables \ref{tab:appdx_dataset_statistics} and \ref{tab:appndx_examples} in the Appendix, respectively.
%The first group is comprised of standard datasets from image clustering research. %community due to their varying complexities.
%Since they are less challenging modern image encoders, we include additional, more challenging, datasets.

\noindent
\textbf{Standard Datasets.} We utilize three widely-used image clustering benchmarking datasets: STL10 \citep{pmlr-v15-coates11a}, Cifar10 \citep{Krizhevsky2009LearningML} and ImageNet10 \citep{imagenet_cvpr09}.

% \noindent
% \textbf{Large Ground Truth Datasets.} In order to evaluate the performance of our method in scenarios with a large number of ground truth clusters, we incorporated Places365 and Tiny Imagenet. Places365 \citep{zhou2017places} comprises a vast collection of images representing various indoor and outdoor scenes, while Tiny Image is a subset of ImageNet.
\noindent
\textbf{Background Datasets.} To assess the robustness of our proposed method against background noise, we include Sports10 \citep{trivedi2021contrastive} and iNaturalist2021 \citep{inaturalist-2021}, two datasets containing high-resolution images of sports scenes in video games and natural environments.

\noindent
\textbf{Human Interpretable Datasets.} Three datasets focusing on human concepts rather than individual objects are included. 
LSUN \citep{yu2015lsun}, showing, e.g., a living room or a kitchen, Human Activity Recognition (HAR) \citep{har-kaggle-22}, containing scenes such as running and Facial Expression Recognition (FER2013) \citep{BarsoumICMI2016}, e.g., surprise, are considered.

% Blip2: 
% -	https://arxiv.org/pdf/2301.12597.pdf

% Vision Encoder:
% -	CLIP Radford et. Al, 21
% o	Vit-L/14: WiT dataset, proprietary OpenAI, 400M image-text pairs
% o	
% -	EVA-CLIP Fang et al., 22 
% o	ViT-g/14: CC12M, COCO, ADE20K, Im21k, Obj365
% o	Seems to be ours
% o	https://arxiv.org/pdf/2211.07636.pdf

% -	Blip2: Keep LM and ViT fixed, train additional QueryTransformer
% o	We have Vit-g/14 and Flan-T5-XL
% o	Flan-T5-XL: 	3B params
% o	Vit-g/14:	1B params
% o	Ours: 		100M trainable params

\section{Text-Guided Image Clustering}

We explore the potential of generated text for image clustering.
First, we use standard image captioning and observe that the text representation outperforms the image representation of several models.
Second, we guide the text generation using VQA models to generate keywords, which we call \textit{keyword-guided clustering}, and introduce \textit{prompt-guided clustering}, where we use domain-specific prompts to elicit relevant properties.
Third, we use the generated text for cluster explainability, obtaining keyword-based descriptions for each cluster.

\subsection{Caption-Guided Image Clustering}
\label{subsec:cgic}

Modern foundation models provide the possibility to work with multiple modalities. 
In particular, image captioning models describe images with text.
Thus, as a first experiment, we investigate how well text clustering on captioned text works in comparison to image clustering, and establish a consistent experimental setup.

\begin{table*}[ht]
\resizebox{.99\textwidth}{!}{%

\begin{tabular}{cc | cccccccccccccccccc}
\toprule
      &       & \multicolumn{6}{c}{Standard} & \multicolumn{4}{c}{Background} & \multicolumn{8}{c}{Human} \\
   Model   &    Representation   & \multicolumn{2}{c}{STL10} & \multicolumn{2}{c}{Cifar10} & \multicolumn{2}{c}{ImageNet10} & \multicolumn{2}{c}{Sports10} & \multicolumn{2}{c}{iNaturalist2021} & \multicolumn{2}{c}{FER2013} & \multicolumn{2}{c}{LSUN} & \multicolumn{2}{c}{HAR} & \multicolumn{2}{c}{Avg} \\
      &       &                         Acc &                         NMI &                         Acc &                         NMI &                        Acc &                         NMI &                         Acc &                         NMI &                         Acc &                        NMI &                         Acc &                        NMI &                         Acc &                         NMI &                         Acc &                         NMI &                         Acc &                         NMI \\
\midrule
Flamingo & Image &                        95.0 &              \textbf{95.13} &                        84.0 &                       84.19 &             \textbf{99.38} &              \textbf{98.85} &              \textbf{75.87} &              \textbf{81.61} &                        40.8 &                      58.09 &              \textbf{36.79} &             \textbf{17.33} &                       60.67 &                       60.98 &                       50.07 &                       43.67 &                       67.82 &              \textbf{67.48} \\
       & TF-IDF &                       82.22 &                        77.0 &                       81.85 &                       76.23 &                      94.32 &                       89.57 &                       54.16 &                       49.86 &                       34.27 &                      43.63 &                       25.77 &                       2.91 &              \textbf{70.58} &                       64.04 &                       40.92 &                       35.52 &                       60.51 &                       54.85 \\
       & SBERT &              \textbf{97.74} &                       94.68 &              \textbf{93.64} &              \textbf{86.15} &                      98.36 &                       96.05 &                       60.32 &                       55.89 &              \textbf{44.93} &             \textbf{58.99} &                       29.79 &                       9.77 &                       68.96 &              \textbf{68.41} &              \textbf{51.37} &              \textbf{46.84} &              \textbf{68.14} &                        64.6 \\
\midrule
GIT & Image &                       51.15 &                       63.62 &                       66.37 &                       64.87 &                      95.41 &              \textbf{93.78} &                       71.17 &                       75.69 &                       42.47 &                       53.0 &                        24.1 &              \textbf{2.15} &                       52.06 &                       51.78 &                       38.81 &                       33.18 &                       55.19 &                       54.76 \\
       & TF-IDF &                       79.92 &                       74.71 &                        74.0 &                       66.73 &                      82.69 &                       76.78 &              \textbf{87.42} &                        84.6 &                       36.12 &                      42.84 &                       25.24 &                       1.66 &                       65.34 &                       57.68 &                       42.87 &                       36.05 &                        61.7 &                       55.13 \\
       & SBERT &              \textbf{96.58} &              \textbf{93.34} &              \textbf{86.79} &              \textbf{76.97} &             \textbf{96.37} &                       92.72 &                       85.73 &              \textbf{88.14} &              \textbf{46.04} &             \textbf{58.78} &              \textbf{26.61} &                       1.95 &              \textbf{69.82} &              \textbf{61.95} &              \textbf{48.11} &              \textbf{42.66} &              \textbf{69.51} &              \textbf{64.56} \\
\midrule
BLIP-2 (*) & Image &  \underline{\textbf{99.65}} &  \underline{\textbf{99.16}} &  \underline{\textbf{98.69}} &  \underline{\textbf{97.59}} &  \underline{\textbf{99.8}} &  \underline{\textbf{99.35}} &                       91.31 &                       93.22 &                       44.97 &  \underline{\textbf{62.7}} &                       35.97 &  \underline{\textbf{21.2}} &                       62.07 &                       64.47 &  \underline{\textbf{52.65}} &  \underline{\textbf{47.06}} &                       73.14 &                       73.09 \\
       & TF-IDF &                        83.3 &                       79.35 &                        89.0 &                       84.75 &                      93.54 &                       88.81 &  \underline{\textbf{99.38}} &  \underline{\textbf{98.65}} &                       34.17 &                      39.07 &                       31.86 &                       6.89 &                       76.69 &                       71.05 &                       50.51 &                       46.09 &                       69.81 &                       64.33 \\
       & SBERT &                       98.03 &                       96.27 &                       97.31 &                       94.07 &                      98.22 &                       96.63 &                       99.07 &                       98.47 &  \underline{\textbf{47.43}} &                      61.63 &  \underline{\textbf{38.21}} &                      20.53 &  \underline{\textbf{81.11}} &  \underline{\textbf{74.37}} &                       50.85 &                       46.68 &  \underline{\textbf{76.28}} &  \underline{\textbf{73.58}} \\
\bottomrule
\end{tabular}´

}
\caption{Comparison of Clustering Accuracy and NMI of image space and generated captions, using TF-IDF and SBERT representations, of multiple Image-to-Text models. 
For each combination of dataset and metric, underlined numbers represent the best overall performance, and bold numbers the best performance per model. (*) Note that BLIP-2 is pre-trained on ImageNet21K \citep{imagenet_cvpr09}, which STL10 and ImageNet10 are subsets of.
}
\label{tab:captioning_results}
\end{table*}

\noindent\textbf{Setup.} 
% General exp. setup / RQ
The commonality between current image captioning models is that they consist of an image encoder and a generative LLM to generate text conditioned on the latent image space.
As described in Section \ref{subsec:exp_setup} we assess the quality of image and generated text by comparing the clustering performance of the vision encoder embeddings with TF-IDF and SBERT representations using K-Means.
We benchmark three SOTA image-to-text models, namely a community-trained version of Flamingo\footnote{\url{https://huggingface.co/dhansmair/flamingo-mini}} \citep{alayrac2022flamingo}, GIT\footnote{\url{https://huggingface.co/microsoft/git-large}} \citep{wang2022git}, and BLIP-2\footnote{\url{https://huggingface.co/Salesforce/blip2-flan-t5-xl}} \citep{li2023blip2}, all available within the Huggingface Transformers library \citep{wolf-etal-2020-transformers}. 
Note that we abstain from including dedicated clustering methods \cite{9878974, Qian_2023_ICCV, gao-etal-2021-simcse} because they are based on a much weaker image encoder, thus achieving much lower performance. Furthermore, it is not straightforward to train transformer-based image models using clustering objectives.
We probabilistically sample a maximum of 80 tokens, without any additional parameters. 
Only for Flamingo, we set top-K to $8$, following the original repository. Experiments were performed on a single A100 40GB and took about 40h hours.
%A more detailed model description is given in Section \ref{subsec:image_text_models}.

% Number of captions
We start by studying the effect of the number of captions generated per image. 
\begin{figure}[t]
\centering
\includegraphics[width=\columnwidth]{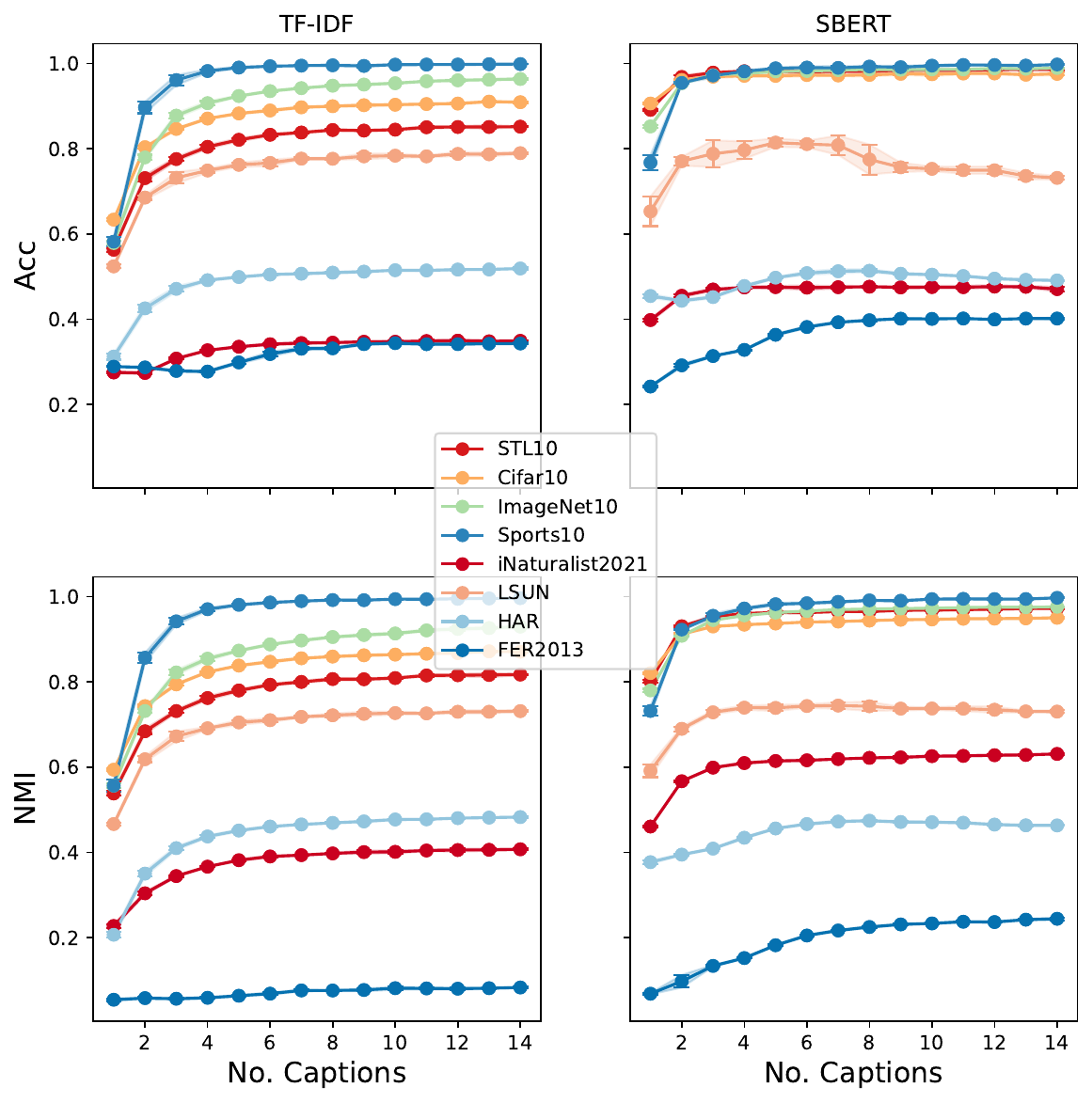}
\caption{Effect of the number of captions sampled per image for BLIP-2. %Investigation of how many captions should be sampled per image. 
The number of captions is depicted on the X-axis, mean and standard deviation of clustering performance are on the Y-axis.}
\label{fig:num_captions}
\end{figure}
For each amount of captions, we sample 6 versions and report the mean and standard error in Figure \ref{fig:num_captions}. 
%To save resources, we only choose four representative datasets across the three groups: Cifar10, Sports10, LSUN, and HAR.

%See appendix \ref{appendix:experiment_descr} for a detailed description of the sampling scheme.

\noindent\textbf{Results.} 
% Captioning analysis
We observe that, for TF-IDF, with a growing number of captions, the performance increases monotonically, whereas SBERT saturates for many datasets. 
Being counting-based, we think that the reason is that TF-IDF is better at reducing the effect of outlier captions, i.e. single bad captions.
For all following experiments, we choose to sample $6$ text generations as a trade-off between sampling efficiency and clustering performance.

%Image captioning
The full image captioning results are shown in Table \ref{tab:captioning_results}.
%SBERT best
The average scores (Avg) show that SBERT outperforms the other two representations across all model types on almost all datasets, while the TF-IDF representation performs worst. 
Note that we abstain from sophisticated preprocessing such as lemmatization or stemming, which is common for frequency-based representations such as TF-IDF, to keep the setup simple and depend on text information as purely as possible. This might (to a certain degree) explain the worse performance.
%For GIT, the Tf-IDF and Image representation perform similarly whereas the image representation of Blip2 is much better than Tf-IDF. 
%Possibly this is because GIT fine-tunes the image encoder with multi-modal data, weakening its unimodal abilities. 

\begin{table*}[ht]
    \centering
    \resizebox{.99\textwidth}{!}{

\begin{tabular}{cc|cccccccccccc}
\toprule
&  & \multicolumn{2}{c}{Sports10} & \multicolumn{2}{c}{iNaturalist2021} & \multicolumn{2}{c}{FER2013} & \multicolumn{2}{c}{LSUN} & \multicolumn{2}{c}{HAR} & \multicolumn{2}{c}{Avg} \\
            & &           Acc &                NMI &           Acc &                NMI &           Acc &                NMI &           Acc &                NMI &           Acc &                NMI &                Acc &                NMI \\
\midrule
Image & ViT & 91.31 & 93.22 & 44.97 & \textbf{62.70} & 35.97 & 21.2 & 62.07 & 64.47 & 52.65 & 47.06 & 57.39 & 57.73 \\
\midrule
Caption-Guided & TF-IDF & \textbf{99.38} & \textbf{98.65} & 34.17 & 39.07 & 31.86 & 6.89 &  \underline{76.69} &  \underline{71.05} & 50.51 & 46.09 &  58.52 &   52.35 \\
& SBERT & 99.07 &  \underline{98.47} & 47.43 &  61.63 & 38.21 & 20.53 & \textbf{81.11} & \textbf{74.37} & 50.85 & 46.68 & 63.33 & 60.34 \\
\midrule
Keyword-Guided & TF-IDF &  \underline{99.08} & 97.82 & 42.13 & 48.25 & \textbf{47.05} & 27.34 & 76.2 &  69.28 & 51.35 &  45.47 & 63.16 & 57.63 \\
& SBERT & 96.89 & 96.87 &  \underline{48.44} & 59.48 & 46.44 & 29.96 & 70.63 & 70.82 & \underline{55.66} &  \underline{50.07} &  \underline{63.61} &  \underline{61.44} \\
\midrule
Prompt-Guided & TF-IDF & 84.83 & 94.46 & 38.01 & 47.61 & \underline{46.86} & \underline{34.25} & 66.4 & 59.92 & 52.74 & 47.96 & 57.77 & 56.84 \\
& SBERT & 98.70 & 98.12 & \textbf{48.57} &  \underline{62.23} & 45.60 & \textbf{36.04} & 71.59 &  63.54 & \textbf{60.93} &  \textbf{52.94} & \textbf{65.08} & \textbf{62.57} \\
\bottomrule
\end{tabular}
}
    \caption{Comparison of clustering performance of the BLIP-2 image encoder features, and examined types of generated text. For prompt-guided clustering, the clusterings belonging to the prompt with the lowest K-Means are evaluated. For each dataset and metric combination, the best performance is bold, and the second-best performance is underlined.
    }
    \label{tab:keywords_vqa_main_table}
\end{table*}

% Blip2 better
Further, we observe that BLIP-2 is the best-performing model. It performs especially well on the standard datasets, which we think is due to the fact that it was pre-trained on ImageNet21k in a self-supervised fashion.

% Conclusion of captioning
In summary, the results show that text representations, obtained only based on (latent) image representations, provide competitive clustering performance, often outperforming the corresponding image representation.
\subsection{Knowledge Injection}
\label{subsec:knowledge_injection}

% Main message: Wir können die Textgenierung auf den Task Image Clustering generell zuschneiden
Now we investigate the potential of guiding the text generation so that it is specifically suited for clustering. 
By using modern VQA models, it is possible to elicit dedicated information from images.
In the following, we introduce two ways to make use of VQA models.

\noindent\textbf{Keyword-Guided Clustering. } 
Given that it is common to (verbally) describe clusters using keywords, we hypothesize that it is beneficial to prompt the model to generate keywords.
The reasons are: 1) keywords provide useful inputs for simpler, traditional count-based representations such as TF-IDF, 2) keywords are useful for count-based analysis methods, such as the proposed cluster explainability algorithm in section \ref{subsec:explainability}, and 3) ground truth cluster labels (as given by classification datasets used in the clustering literature) are typically described using only a few keywords.

\noindent\textbf{Prompt-Guided Clustering. } In real-world scenarios, often, some domain knowledge about the given data is available. 
The ability of VQA models to retrieve dedicated information from images opens up the possibility of using domain knowledge in the natural form of text. An example is to ask "Which activity is performed in the picture?". 
Note, crucially, that this is not possible using standard image clustering models. 
%We refer to this as \textit{Prompt-Guided Clustering}.

\noindent\textbf{Setup. } 
%Based on the results of the last section, we use Blip-2 and sample 4 texts for each image. 
Due to resource constraints, we only use the best-performing (cf. Table \ref{tab:captioning_results}) image-to-text model, BLIP-2, for the subsequent experiments.
% Also, BLIP-2 allows for a fairer comparison because the text and image sub-modules are fixed and not fine-tuned with multi-modal data.
Based on the results depicted in Figure \ref{fig:num_captions}, we sample $k=6$ texts for each image.

% As explained previously, BLIP-2 solves the standard datasets, achieving a performance close to 100\% (not shown), they are not included in the following discussion. 
% Additionally, they only exhibit a collection of objects, rendering it difficult to inject domain knowledge other than `What objects are described?'.
% Keyword guided
For keyword-guided clustering, we use the question "Which keywords describe the image?".
% VQA
To perform prompt-guided clustering, we create four questions for each of the datasets.
The questions were created by naively transforming the name of the dataset into a question, e.g. for human action recognition "Which activity is performed?" is asked. Note, that no additional prompt engineering efforts were made, as we are not aware of a more principled way to design such prompts.
Find all questions in Table \ref{tab:full_questions} in Appendix \ref{appdx:sec_ded_vqa}. 

%Datasets
BLIP-2 solves the ``standard'' datasets with almost 100\% and they exhibit only a collection of objects, making it difficult to pose interesting questions other than `What objects are described?'. Thus, they are excluded in the following experiments.
% Also, BLIP-2 solves the standard datasets, achieving a performance close to 100\%, making them less interesting. 
% K-Means loss
It is well known that current LLMs possibly generate very different texts, even though the prompt has the same meaning \cite{elazar-etal-2021-measuring}.
Therefore, in Table \ref{tab:keywords_vqa_main_table} we use an unsupervised heuristic to decide which prompt works best by taking the prompt belonging to the clustering with the lowest K-Means loss.

% \begin{figure}[h]
% \centering
% \includegraphics[width=0.35\textwidth]{aux/plots/scatter_seaborn.pdf}
% \caption{Comparison of all prompts with previous BLIP2 representations.
% We observe that dedicated questions can work well but are prone to large differences in performance.}
% \label{fig:vqa_ded_questions}
% \end{figure}

\begin{table}%[h]
\centering
\resizebox{.48\textwidth}{!}{%

\begin{tabular}{l | cc}
\toprule
 Modality / Question &  \multicolumn{2}{c}{SBERT} \\
 &    Acc &    NMI \\
\midrule
Image                                           &  52.65 &  47.06 \\
Which keywords describe the image?              &  55.66 &  50.07 \\
What type of motion is depicted in the picture? &  49.20 &  42.54 \\
Which activity is shown in the picture?         &  56.03 &  49.69 \\
Which action is shown in the picture?           &  58.68 &  52.86 \\
What is the person doing in the picture?        &  \textbf{60.93} &  \textbf{52.94} \\
\bottomrule
\end{tabular}

}
\caption{A case study for prompt-guided image clustering on Human Action Recognition, using the SBERT representation. Find the full table in Appendix \ref{appdx:sec_ded_vqa}.}
\label{tab:ded_vqa_example}
\end{table}

% \begin{figure}
%     \centering
%     \includegraphics[width=0.5\textwidth]{aux/plots/num_kws_fig_sbert_accuracy.pdf}
%     \caption{An investigation of how many keywords are needed. The lines show the mean and standard error of the clustering accuracy of the SBERT representation.} %TODO: other plots in Appendix
%     \label{fig:num_kws}
% \end{figure}

\begin{figure*}[ht]
    \centering
    \includegraphics[width=.99\textwidth]{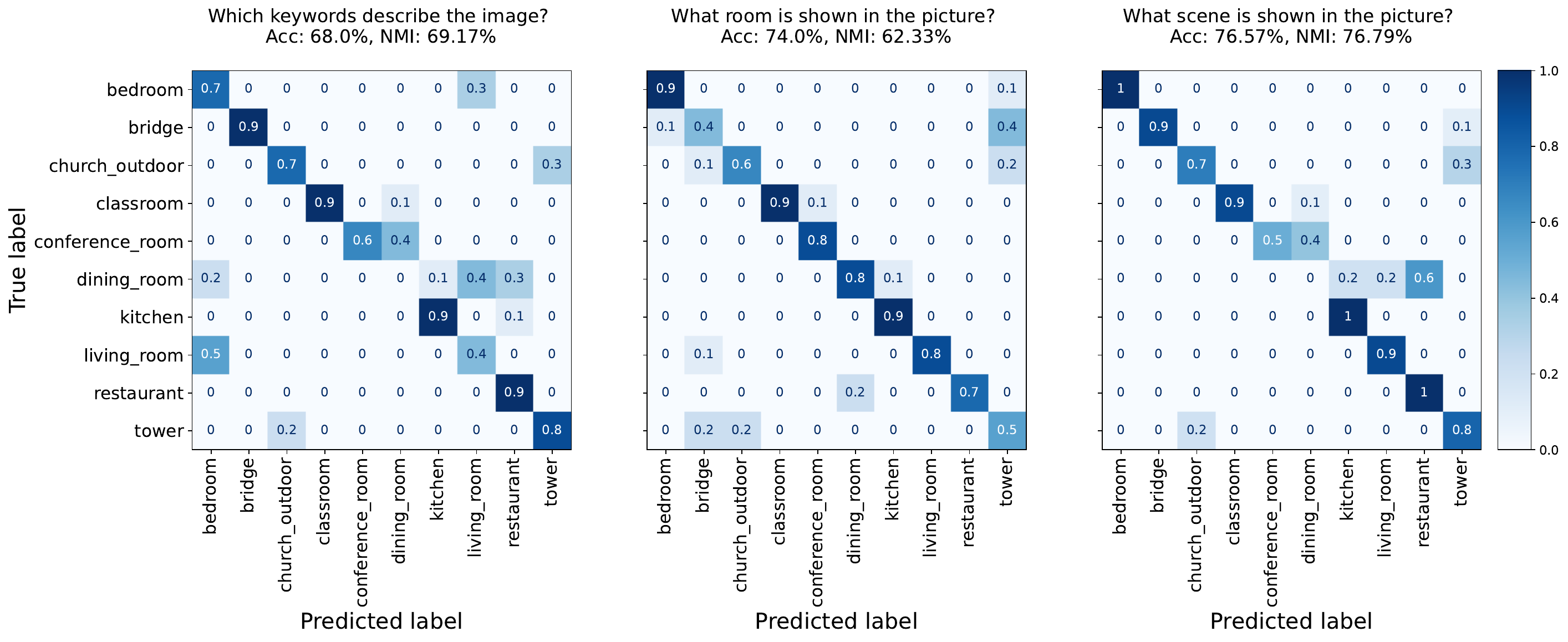}
    \caption{Confusion matrices based on three clustering results from text generated with three different VQA prompts. While a similar cluster accuracy is achieved, we observe that the clustering relates to the prompt. In the middle all room clusters are clustered well, on the right side the clustering is not able to distinguish well between dining room, kitchen and restaurant (see corresponding dining room row), but leads to better overall accuracy.}
    \label{fig:vqa_confusions}
\end{figure*}

\noindent\textbf{Results.} 
%Table \ref{tab:keywords_vqa_main_table} compares the image backbone to the generated text and the generated keywords, named "Image", "Captions" and "Keywords" in the top column.
In Table \ref{tab:keywords_vqa_main_table} we observe that the average performance (Avg) for caption-guided image clustering and SBERT-based keyword-guided clustering is similar.
Using keywords, TF-IDF improves on average by 5\% for both cluster accuracy and NMI, closing the gap to SBERT.
This result is in line with our hypothesis that keywords are a useful representation for image clustering.

% Numbers improved; there is variance
As a case study, Table \ref{tab:ded_vqa_example} holds the results for the HAR dataset.
We observe a notable variance in the performance of multiple prompts.
This is a common phenomenon for prompting-based methods \citep{pmlr-v139-zhao21c}.
Using the K-Means loss as a proxy for selecting the best prompt leads to the best average performance in Table \ref{tab:keywords_vqa_main_table}.

% Confusion matrix: 
Interestingly, the confusion matrices in Figure \ref{fig:vqa_confusions} show different assignment patterns depending on the question posed to the VQA model.
For instance, when posing the question `What room is shown in the picture?', all room clusters are formed well, but the others, e.g. bridge or tower, are worse.
We argue that this variation is not an issue but a feature of prompt-guided image clustering, e.g., during exploratory data analysis, where one might want to investigate different aspects of a dataset. 

In summary, we demonstrate that it is possible to improve clustering performance by injecting domain knowledge in the form of text and that the clustering changes according to the posed questions. Further examples of the impact of different prompts on the embedded space and clustering are shown in t-SNE embeddings in Figures \ref{fig:image_caption_classroom_conferenceroom} and \ref{fig:image_caption_carracing_bikeracing} in the Appendix.

\subsection{Cluster Explainability}
\label{subsec:explainability}

% Wir können Cluster in der natürlichsten Art und Weise beschreiben: Mit Text
So far, we use the generated text solely to form clusters.
But given the (built-in) interpretability of text, a natural extension is to use text as an explanation of the formed clusters.
Explainability for image clustering is an important issue, as it provides insights into how the clustering algorithm groups the images, helping users understand the underlying patterns and relationships.
%Additionally, cluster descriptions provide the possibility to validate the results of the algorithm and ensure that the clusters it has identified are meaningful and relevant.
The availability of textual descriptions for each cluster sample allows us to extrapolate to textual descriptions of each cluster as a whole.%, improving the explainability of the clustering. 
Note that this is not possible using models considering only images.

%As previously mentioned, the used datasets are naturally geared towards classification, so the natural description is given by keywords. 
We hypothesize that a concise way to describe a cluster is to use a small set of keywords. 
This is based on the fact that the considered datasets use keyword-based labels.
Thus, we introduce the following algorithm to obtain keywords for each cluster from the generated text.

\noindent\textbf{Explainability Algorithm. } 
For each predicted cluster, the keywords are sorted by their number of occurrences in the generated texts.
The algorithm returns the most frequent keywords per cluster. 
If a keyword occurs in multiple cluster descriptions, it is not considered, and the next most occurring is chosen.
We take the two most occurring keywords based on an initial screening of the LSUN dataset.
Find the Pseudocode in the Appendix \ref{appendix_sec:explainability}.%in Algorithm \ref{alg:xai_algorithm}.\\

\noindent\textbf{Setup.} 
We provide a quantitative analysis of the generated descriptions by applying two metrics. 
First, we introduce the subset exact match (SEM) metric, for which we lowercase each string and check whether the ground truth cluster name appears in the predicted keywords.
No further standardization, such as stemming or lemmatization, is performed.
Second, SBERT embeddings are used to check the similarity between cluster names and keywords obtained by the explainability algorithm. 
According to our initial investigation, we use a cosine similarity of $0.4$ as the threshold to indicate a match between ground truth and explanation.
For each dataset, we provide the cluster accuracy and the explainability performance given the ground truth (\textit{Truth}) clustering and the predicted (\textit{Pred}) clustering, corresponding to the cluster accuracy.
Out of the 50 conducted K-Means runs, we use the clustering with the lowest K-Means loss for the analysis.
\begin{table}%[h]
\centering
\resizebox{.48\textwidth}{!}{%

% \begin{tabular}{ll|ll}
% \toprule
%         \multicolumn{2}{l}{Sports10} & \multicolumn{2}{l}{LSUN} \\
%               GT &                         XAI &              GT &                     XAI \\
% \midrule
% AmericanFootball &            football, player &         bedroom &            bedroom, bed \\
%       Basketball & basketball, basketball game &          bridge &           bridge, river \\
%       BikeRacing &            race, motorcycle &  church\_outdoor &       church, cathedral \\
%        CarRacing &                 car, racing &       classroom &      classroom, teacher \\
%         Fighting &               fight, boxing & conference\_room &     meeting, conference \\
%           Hockey &                hockey, puck &     dining\_room & dining table, furniture \\
%           Soccer &         soccer, soccer game &         kitchen &            wood, island \\
%      TableTennis &     ping pong, table tennis &     living\_room &     living room, living \\
%           Tennis &         tennis, tennis game &      restaurant &         restaurant, bar \\
%       Volleyball &           volleyball, beach &           tower &             tower, city \\
% \bottomrule
% \end{tabular}

\begin{tabular}{l | lcc}
\toprule
         Ground Truth &              Explanation & SEM & SBERT Sim. \\
\midrule
Sports10 & & &\\
 \midrule
% AmericanFootball &               football, nfl &   1 &          1 \\
%       Basketball & basketball, basketball game &   1 &          1 \\
%       BikeRacing &           motorcycle, rider &   1 &          1 \\
%        CarRacing &                  car, speed &   0 &          0 \\
%         Fighting &               fight, boxing &   1 &          1 \\
%           Hockey &         hockey, hockey game &   1 &          1 \\
%           Soccer &         soccer, soccer game &   1 &          1 \\
%      TableTennis &     ping pong, table tennis &   0 &          0 \\
%           Tennis &         tennis, tennis game &   1 &          1 \\
%       Volleyball &           volleyball, beach &   1 &          1 \\
AmericanFootball &               football, nfl &   0 &          1 \\
      Basketball & basketball, basketball game &   1 &          1 \\
      BikeRacing &           motorcycle, rider &   0 &          1 \\
       CarRacing &                  car, speed &   0 &          0 \\
        Fighting &               fight, boxing &   0 &          1 \\
          Hockey &         hockey, hockey game &   1 &          1 \\
          Soccer &         soccer, soccer game &   1 &          1 \\
     TableTennis &     ping pong, table tennis &   0 &          0 \\
          Tennis &         tennis, tennis game &   1 &          1 \\
      Volleyball &           volleyball, beach &   1 &          1 \\
\midrule
LSUN & & &\\
\midrule
%         bedroom &              bedroom, bed &   1 &          1 \\
%          bridge &             bridge, river &   1 &          1 \\
%  church\_outdoor &         church, cathedral &   1 &          1 \\
%       classroom &        classroom, teacher &   1 &          1 \\
% conference\_room &       meeting, conference &   1 &          1 \\
%     dining\_room & dining room, dining table &   1 &          1 \\
%         kitchen &             kitchen, wood &   1 &          1 \\
%     living\_room &       living room, living &   1 &          1 \\
%      restaurant &           restaurant, bar &   1 &          1 \\
%           tower &               tower, city &   1 &          1 \\
        bedroom &              bedroom, bed &   1 &          1 \\
         bridge &             bridge, river &   1 &          1 \\
 church\_outdoor &         church, cathedral &   0 &          1 \\
      classroom &        classroom, teacher &   1 &          1 \\
conference\_room &       meeting, conference &   0 &          1 \\
    dining\_room & dining room, dining table &   1 &          1 \\
        kitchen &             kitchen, wood &   1 &          1 \\
    living\_room &       living room, living &   1 &          1 \\
     restaurant &           restaurant, bar &   1 &          1 \\
          tower &               tower, city &   1 &          1 \\
\bottomrule
\end{tabular}

}
\caption{Examples of generated explanations for Sports10 and LSUN.  If a value in the SEM or SBERT Sim. column is 1, it means that the metric says ground truth and explanation match.
}
\label{tab:xai_example}
\end{table}

\noindent\textbf{Results.} 
Table \ref{tab:xai_performance} depicts the quantitative evaluation of our algorithm.
We observe that the SBERT metric is always equal to or higher than the SEM metric, which makes sense as SEM is a rather strict metric, not understanding synonyms or syntactical changes, e.g., "TableTennis" vs. "table tennis".
Interestingly, in most cases, the SBERT metric is higher than the clustering accuracy.
Table \ref{tab:xai_example} shows an example of generated descriptions and metrics. 
We observe that both metrics cannot understand that ``TableTennis'' and ``ping pong, table tennis'' have the same meaning, but still, all cluster descriptions of Sports10 are correct. 
For iNaturalist2021 and FER2013, we observe that the generated text is often of bad quality, resulting in low-quality descriptions.
We conclude that the generated descriptions provide a good overview of the content of the generated clusters and in most cases, describe the dataset better than clustering accuracy suggests.
\section{Broader Impact}
\label{ref:broader_impact}
% Or call this discussion?

We believe there is a lot of unused potential for text as an abstraction in image clustering. 
%In the following, we discuss two topics.

% Clustering always talks about meaningful, text provides good opportunities
\noindent\textbf{Text as a proxy for ``meaningful'' clustering.}
Clustering research aims to find meaningful clusters. 
In general, it is an open question to define what meaningful exactly stands for, some researchers even call it an ill-posed problem.
We argue that text is a good proxy to express meaningfulness as it is based on the natural human form of communication.
This is a novel viewpoint on the task of image clustering aligning with research methodologies in the clustering community, where clustering methods are commonly benchmarked with datasets that have human-annotated textual labels as ground truth.
%This is also backed by decades of clustering research because it typically assumes human-given textual labels as ground truth.
Our research contributes to the discussion about meaningful clustering by showing that generated text improves the interpretability of the detected clusters. 

\noindent\textbf{Knowledge Injection. }
% Zero-Shot learning
% Put explainability back into clustering
Furthermore, it can be highly subjective what determines a meaningful clustering.
For a given dataset, different people are interested in different types of information. 
For example, in real-world scenarios, an expert might have several questions about a dataset based on their domain knowledge. 
We show that these questions can be used to guide the clustering process by prompting VQA models.
%We show that this domain knowledge can be included in the clustering process via prompting VQA models.
%, and that it is beneficial.
% Based on this subjectivity, in real-world scenarios, often a high level understanding of the dataset or of the information someone is interested in, i.e. domain knowledge, is available.
%We have shown that it is beneficial to include this knowledge via prompts.
Given the current speed of research, we believe that the increasing ability to use more detailed prompts will drastically improve our knowledge injection method. 
This, in turn, will open up new research avenues for injecting knowledge into the clustering process.

\begin{table}%[h]
\centering
\resizebox{.48\textwidth}{!}{%

\begin{tabular}{l |cc | cc | cc}
\toprule
% {} &  & \multicolumn{2}{c}{Fuzzy Match} & \multicolumn{2}{c}{SBERT Sim.} \\
% {} &  cluster acc &       Truth &            Pred &      Truth &            Pred \\
% Dataset         &              &             &                 &            &                 \\
% \midrule
% STL10           &        87.07 &      100.00 &  \textbf{100.0} &     100.00 &  \textbf{100.0} \\
% ImageNet10      &        92.85 &       90.00 &            90.0 &     100.00 &  \textbf{100.0} \\
% CIFAR10         &        90.51 &       90.00 &            90.0 &     100.00 &  \textbf{100.0} \\
% \hline
% Sports10        &        98.33 &       90.00 &            90.0 &      80.00 &            80.0 \\
% iNaturalist2021 &        39.27 &       27.27 &            9.09 &      90.91 &  \textbf{45.45} \\
% \hline
% FER2013         &        44.98 &       25.00 &            37.5 &      50.00 &            25.0 \\
% LSUN            &        76.03 &       90.00 &   \textbf{80.0} &     100.00 &  \textbf{100.0} \\
% HAR             &        51.54 &       66.67 &  \textbf{53.33} &      93.33 &   \textbf{80.0} \\

 & \multicolumn{2}{c}{Cluster Acc} & \multicolumn{2}{c}{SEM} & \multicolumn{2}{c}{SBERT Sim.} \\
 &  TF-IDF &  SBERT & Truth &          Pred &      Truth &          Pred \\
% \midrule
% STL10           &      87 &     98 &   100 &  \textbf{100} &        100 &  \textbf{100} \\
% ImageNet10      &      94 &     99 &    90 &            90 &        100 &  \textbf{100} \\
% CIFAR10         &      91 &     97 &    90 &            90 &        100 &  \textbf{100} \\
% Sports10        &      99 &     98 &    80 &            80 &         80 &            80 \\
% iNaturalist2021 &      40 &     48 &    27 &             9 &         91 &            45 \\
% LSUN            &      75 &     68 &    90 &  \textbf{100} &        100 &  \textbf{100} \\
% HAR             &      51 &     56 &    67 &   \textbf{60} &         87 &   \textbf{87} \\
% FER2013         &      46 &     46 &    25 &            25 &         38 &            25 \\

% \bottomrule
\midrule
STL10           &      87 &     98 &   100 &  \textbf{100} &        100 &  \textbf{100} \\
ImageNet10      &      94 &     99 &    30 &            30 &        100 &  \textbf{100} \\
CIFAR10         &      91 &     97 &    90 &            90 &        100 &  \textbf{100} \\
Sports10        &      99 &     98 &    50 &            50 &         80 &            80 \\
iNaturalist2021 &      40 &     48 &     0 &             0 &         91 &            45 \\
LSUN            &      75 &     68 &    70 &   \textbf{80} &        100 &  \textbf{100} \\
HAR             &      51 &     56 &    20 &            13 &         87 &   \textbf{87} \\
FER2013         &      46 &     46 &    12 &            12 &         38 &            25 \\
\bottomrule

\end{tabular}
}
\caption{Evaluation of our explainability method. 
%SEM asks whether the ground truth label is a substring of the explanation, and SBERT similarity if ground truth and prediction pass the cosine similarity threshold $0.4$. 
In ``Truth'', the explainability method is applied to the ground truth clustering whereas in ``Pred'' it is applied to the clustering of the given clustering accuracy.
Numbers are boldened if the explainability score of a found clustering (``Pred'' columns) outperforms clustering accuracies.
}
\label{tab:xai_performance}
\end{table}
\section{Conclusion}

% We introduce, and challenge
In this work, we introduce \textit{Text-Guided Image Clustering}, using image-captioning and VQA models to automatically generate text, and subsequently cluster only the generated text.
% We challenge the current research in image clustering by introducing generated text as a valuable source of information.
% Experiments
After applying multiple captioning models on eight diverse datasets, our experiments show that representations of generated text outperform image representations on many datasets.
% Prompt-Gudining
Further, we use text to include task- and domain knowledge by prompting VQA models, resulting in additional improvements in clustering performance. We find that it is possible to shape the clustering favorably according to the information given by a specific prompt. 
% XAI: TODO
Additionally, we use the generated text to obtain a keyword-based description for each cluster and show their usefulness quantitatively and qualitatively.

% Relation back to the very beginning of the document
%We challenge the current research in image clustering by proposing generated text as an abstraction %that is possibly better suited 
%for image clustering. 

%\noindent Other areas, such as psychology or neuroscience, research the relationship between language and visual information, e.g. by examining how kids understand scenes with or without additional descriptions. In image clustering, research on the potential of more abstract text descriptions is underrepresented. Therefore, we would like to initiate this research with the proposal of using generated text for clustering images. 

\noindent
While it is difficult to identify background noise or irrelevant features in the pixel space, text is discrete and interpretable. We show that text-guided image clustering often outperforms clustering purely on image information, and provides interpretability. Therefore, our research provides insights into the role of text in determining meaningful clusterings.

%In the field of image clustering, research about the possibilities the abstraction of text provides to partition data into meaningful groups is underrepresented.
%We propose to make use of generated text.
% While other areas, such as psychology or neuroscience, research the relationship between language and visual information in human understanding, r

% in depth, research about how language is able to determine a good, or meaningful, image clustering, is underrepresented.
% We propose generated text as %a proxy and 
% a starting point.
\section{Limitations}

% - Only limited for cluster explainability

While our proposed approach shows promising results, several limitations apply.

Text-guided image clustering is dependent on the quality and effectiveness of the generated text. 
In cases where the generated text is incomplete, misleading, or fails to capture the essential features of the images, the clustering algorithm may struggle to accurately group similar images.
Current image-to-text models are mostly trained on data obtained from the internet. 
For example, because of licensing and other restrictions, many domain-specific images are not represented appropriately in the training data, resulting in poor text generation abilities for those domains. 
Nevertheless, our experiments are performed on a wide variety of datasets, more diverse than in common image clustering research, proving the general applicability of the method.
 
While we show that our approach is effective for image clustering, we do not include results for other visual modalities, such as video or 3D point clouds. We show that it is worthwhile to investigate the possibility of clustering images using generated text and generating textual cluster explanations. The rapid advancement of machine learning models will also enable the same approach for other modalities.

%Currently, our focus lies solely on the comparison of images and generated text for the purpose of clustering. We did not explore the potential benefits of combining images and corresponding generated text in the clustering process. The field of multi-view clustering combines multiple heterogenous modalities of data instances into a single clustering. However, multi-view clustering assumes the availability of accurate and reliable data. In order to bridge the gap between the noisy nature of the generated text and the application of multi-view clustering, dedicated research and development efforts are necessary. 

The approach of prompt-guided image clustering is based on the assumption that domain knowledge is readily accessible, allowing the generation of specific questions to guide VQA models. 
While we show that leveraging domain knowledge can prove advantageous, clustering methods are frequently employed for exploratory data analysis. 
Introducing domain knowledge may limit the discovery of novel insights or alternative interpretations due to biased prompts. %that could have emerged from a more open and unbiased clustering approach.

%In practical scenarios, all limitations due to the task clustering, also apply to text-guided image clustering. For instance, without ground truth, it is difficult to evaluate the resulting clusters.

% We have:

% - quality of image-to-text
% - other modalities
% - prompt-guided: availability of domain knowledge - can be good or bad
% - multi-modal clustering

% Other ideas:
% - is img. encoder and decoded text fair comparison? Is trained in a way to work well?
%     dangerous but might take air off reviewer
% - more diverse datasets: just a space filler maybe
% - other clustering than K-Means; other repr. than TF-IFD; SBERt
% - Explainability: There could be more 

% - Maybe good: In all approaches we tried to stay very basic w.r.t methods, but we could increase complexity. E.g. more complex summarization algorithms than count-based

\section*{Acknowledgements}
We thank the anonymous reviewers for their constructive feedback. This research has been funded by the Vienna Science and Technology Fund (WWTF)[10.47379/VRG19008] “Knowledge-infused Deep Learning for Natural Language Processing”. 
We thank the European High Performance Computing initiative for providing the computational resources that enabled this work. EHPC-DEV-2022D10-051, EHPC-DEV-2023D11-017.

\bibliography{custom}%,rolling_review_files/anthology}
\bibliographystyle{rolling_review_files/acl_natbib}

\appendix
\label{sec:appendix}

% \section{Experimental Setup}
% \label{appendix:experiment_descr}

% - extended dataset description
% - describe metrics: Cluster Accuracy / NMI

\section{Dataset Description}
\label{sec:appendix_data}

Here, we provide some additional information about the datasets.
An overview of the datasets is given in Table \ref{tab:appdx_dataset_statistics}, including name, number of classes, number of images, and size, given in pixels.  You can find examples of images of each dataset in Table \ref{tab:appndx_examples}.

In the following, there is a small description of the datasets, including the class labels, provided in their original form which we also use in the evaluation of our explainability algorithm.

\textbf{STL10 \citep{pmlr-v15-coates11a}.} This traditional dataset consists of 10 classes, namely ``deer, horse, bird, cat, ship, airplane, car, truck, monkey, dog''. We use the full dataset, i.e. train and test split. Note, that it is inspired by Cifar10 and attempts to be more complicated because it contains fewer images.

\textbf{Cifar10 \cite{Krizhevsky2009LearningML}.}  The dataset is comprised of 10 similar object classes: ``deer, horse, bird, automobile, airplane, cat, ship, truck, dog, frog''. Again, we use the full dataset.

\textbf{ImageNet10.} Imagenet-10 is a subset of the larger ImageNet dataset, containing 10 classes. Given the hierarchical nature of of ImageNet, each class is described by multiple keywords: 'trailer truck, tractor trailer, trucking rig, rig, articulated lorry, semi', 'snow leopard, ounce, Panthera uncia', 'airliner', 'Maltese dog, Maltese terrier, Maltese', 'sports car, sport car', 'orange', 'soccer ball', 'airship, dirigible', 'container ship, containership, container vessel', 'king penguin, Aptenodytes patagonica'

\textbf{Sports10 \citep{trivedi2021contrastive}.} The Sports-10 dataset provides labeled images from 175 video games across 10 sports genres. The labels are ``CarRacing, Tennis, AmericanFootball, BikeRacing, TableTennis, Fighting, Basketball, Hockey, Soccer, Volleyball''.

\textbf{Inaturalist2021 \citep{inaturalist-2021}.} The full dataset contains images of 10,000 species separated into 10 classes, which are ``Animalia, Arachnids, Amphibians, Birds, Insects, Ray-finned Fishes, Plants, Mollusks, Reptiles, Fungi, Mammals''. We experiment with the validation set.

\begin{table}[ht]
    \centering
    \resizebox{.45\textwidth}{!}{
\begin{tabular}{ll | c|l|l}
\toprule
Dataset Group & Name & No. of classes & No. of Images & Size (pixels) \\
\midrule
Standard & STL10 & 10 & 13000 & 96x96 \\
 & ImageNet10 & 10 & 13000 & 500x364 \\
 & CIFAR10 & 10 & 60000 & 32x32 \\
Background & Sports10 & 10 & 3000 & 1280x720 \\
 & iNaturalist 2021 & 11 & 100000 & 284x222 \\
Human & LSUN & 10 & 3000 & 341x256 \\
 & Human Action Recognition & 15 & 18000 & 240x160 \\
 & FER2013 & 8 & 35488 & 48x48 \\
\bottomrule
\end{tabular}
}
\caption{Overview over some basic dataset statistics.}
\label{tab:appdx_dataset_statistics}
\end{table}

\textbf{LSUN \citep{yu2015lsun}.} The Large-Scale Scene Understanding (LSUN) dataset offers labeled images depicting scenes from the following categories: ``conference\_room, dining\_room, bedroom, church\_outdoor, bridge, tower, restaurant, living\_room, classroom, kitchen''. We experiment with the test set.

\textbf{HAR \citep{har-kaggle-22}.} contains images of human activities. They are ``running, sleeping, listening\_to\_music, texting, drinking, clapping, fighting, eating, sitting, using\_laptop, cycling, calling, laughing, hugging, dancing''.

\textbf{FER2013 \citep{BarsoumICMI2016}.} The Facial Expression Recognition 2013 dataset consists of labeled grayscale images depicting human facial expressions, which are ``surprise, anger, contempt, happiness, fear, disgust, sadness, neutral''.

\begin{table*}[ht]
\centering
\small
% \resizebox{.75\textwidth}{!}{
    \begin{tabular}{l|clcl}
    \toprule
    Dataset  & Image1 & Label1 & Image2 & Label2 \\ 
    \midrule
STL10 & \includegraphics[width=0.1\textwidth]{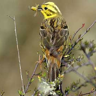}& bird & \includegraphics[width=0.1\textwidth]{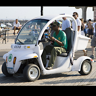}& car \\ 
\midrule \\ 
CIFAR10 & \includegraphics[width=0.1\textwidth]{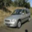}& automobile & \includegraphics[width=0.1\textwidth]{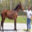}& horse \\ 
\midrule \\ 
ImageNet10 & \includegraphics[width=0.1\textwidth]{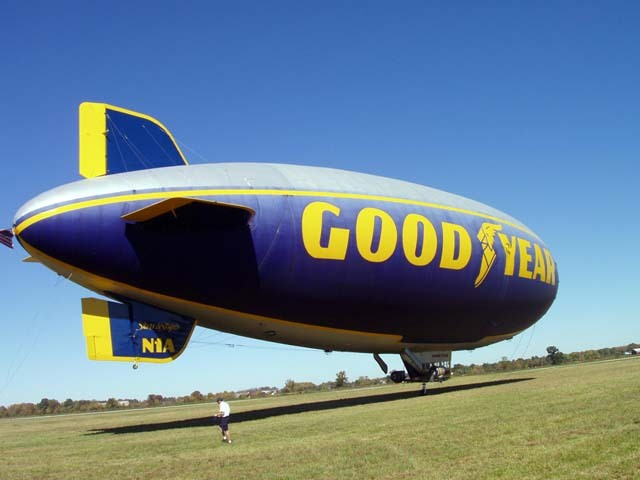}& airship, dirigible & \includegraphics[width=0.1\textwidth]{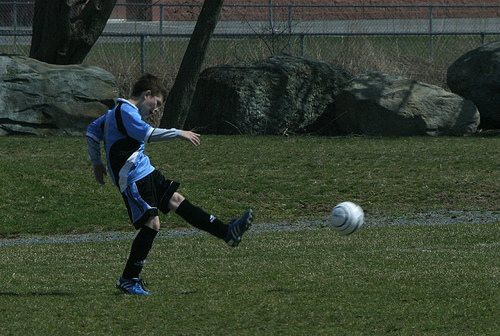}& soccer ball \\ 
\midrule \\ 
Sports10 & \includegraphics[width=0.1\textwidth]{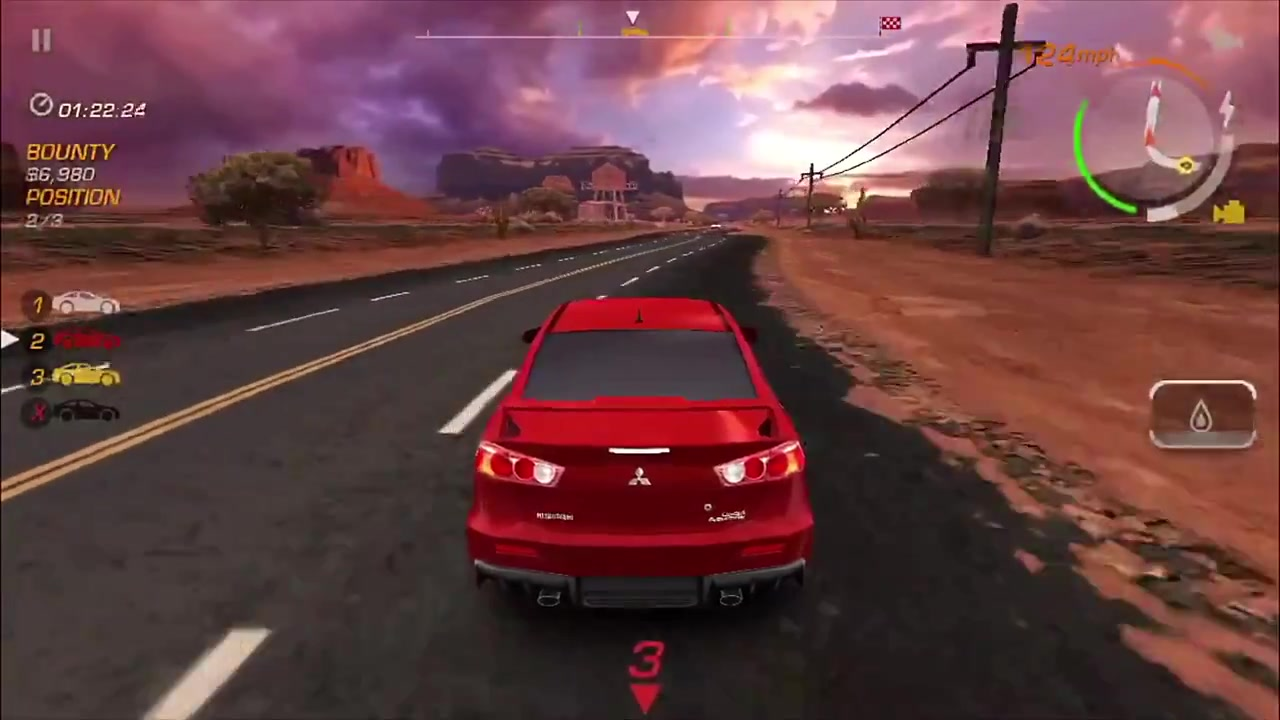}& CarRacing & \includegraphics[width=0.1\textwidth]{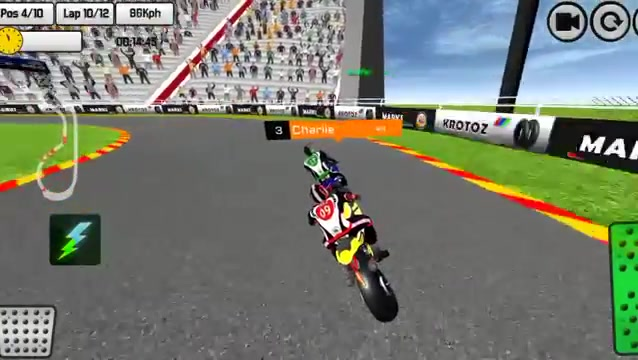}& BikeRacing \\ 
\midrule \\ 
iNaturalist2021 & \includegraphics[width=0.1\textwidth]{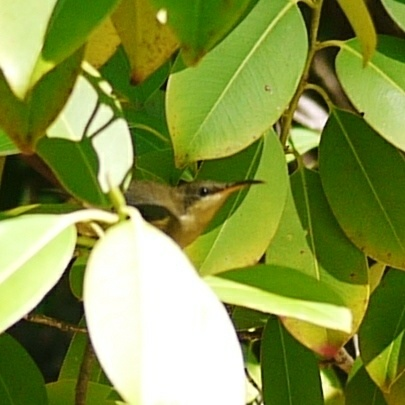}& Birds & \includegraphics[width=0.1\textwidth]{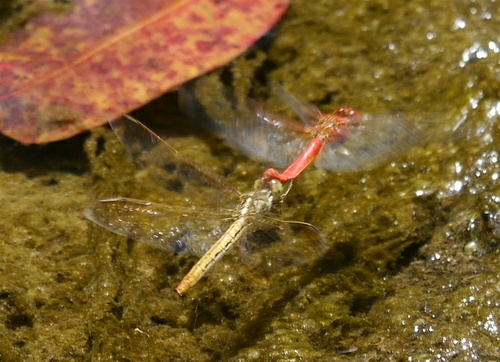}& Insects \\ 
\midrule \\ 
LSUN & \includegraphics[width=0.1\textwidth]{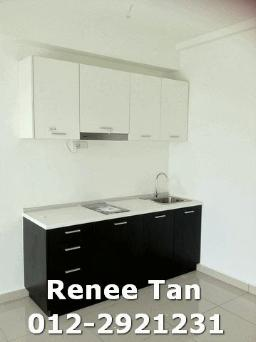}& kitchen & \includegraphics[width=0.1\textwidth]{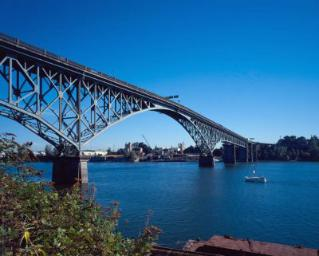}& bridge \\ 
\midrule \\ 
Human Action Recognition & \includegraphics[width=0.1\textwidth]{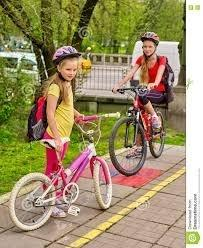}& cycling & \includegraphics[width=0.1\textwidth]{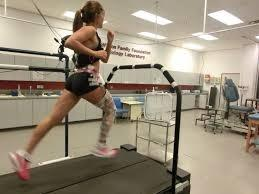}& running \\ 
\midrule \\ 
FER2013 & \includegraphics[width=0.1\textwidth]{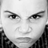}& anger & \includegraphics[width=0.1\textwidth]{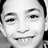}& happiness \\ 
\bottomrule

    \end{tabular}
    % }
    \caption{Examplatory images of the datasets. The images contain different properties, such as image quality or background noise. Also, the labels vary in their syntax and semantic meaning, e.g. objects vs. movements.}
    \label{tab:appndx_examples}
\end{table*}
\section{Knowledge Injection}
\label{appdx:sec_ded_vqa}

\begin{figure*}[ht]
    \centering
    \includegraphics[width=.95\textwidth]{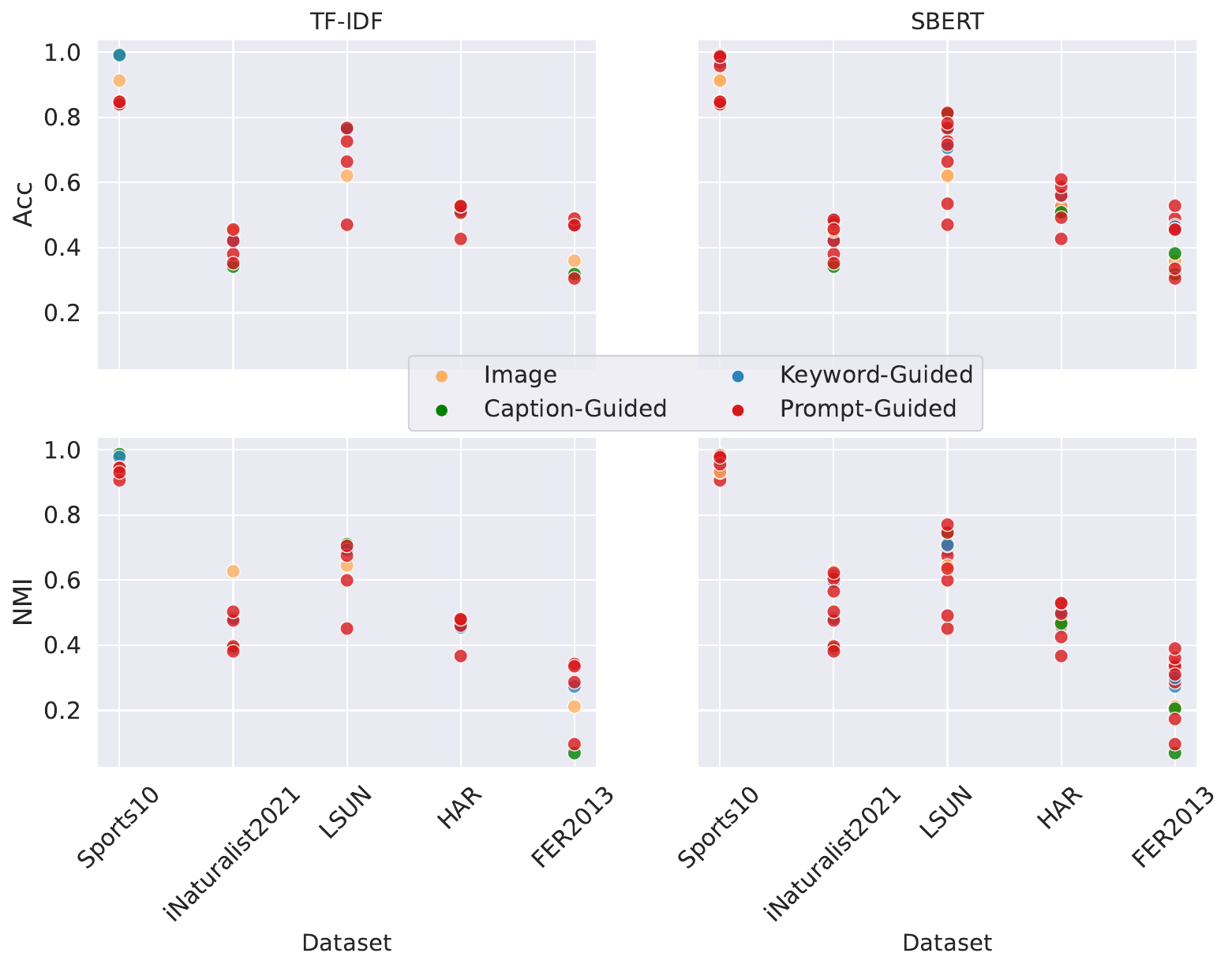}
    \caption{Comparison of all used strategies. Find the questions for prompt-guided clustering in Table \ref{tab:full_questions}.}
    \label{fig:appndx_prompt_performance}
\end{figure*}

In section \ref{subsec:knowledge_injection} we introduce prompt-guided clustering. 
For each dataset, multiple prompts are tested.
They are generated by adapting the dataset name and transforming them into a question.
Table \ref{tab:full_questions} encompasses all prompts used in our experimental setup, accompanied by the corresponding evaluation performance metrics, namely Cluster Accuracy (Acc) and Normalized Mutual Information (NMI) for the image encoder representation, and the TF-IDF and SBERT representations. 
The used model is BLIP-2.
Further, we provide a visual inspection of the same numbers in Figure \ref{fig:appndx_prompt_performance}.

\begin{figure*}[ht]
    \centering
    \includegraphics[width=.95\textwidth]{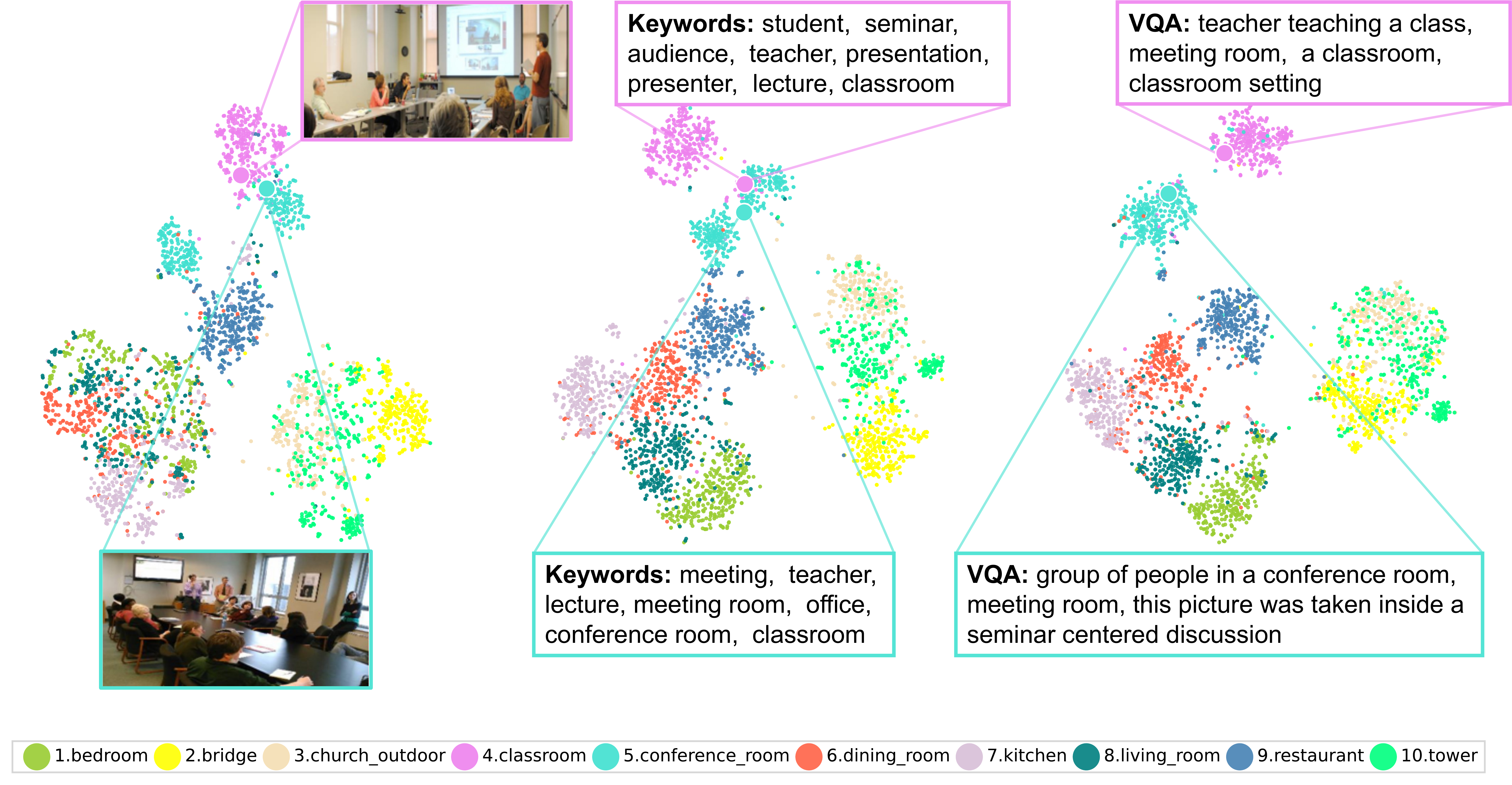}
    \caption{t-SNE embeddings of BLIP2 for the LSUN dataset. From left to right: Image embedding (Acc: 63.11), Keyword SBERT embedding (Acc: 71.12) and VQA SBERT embedding (Acc: 81.83 with prompt: ``What environment is shown in the picture?''). The improvement in cluster accuracy corresponds to better separated clusters in the t-SNE embeddings. }\label{fig:image_caption_classroom_conferenceroom}
\end{figure*}

\begin{figure*}[ht]
    \centering
    \includegraphics[width=.95\textwidth]{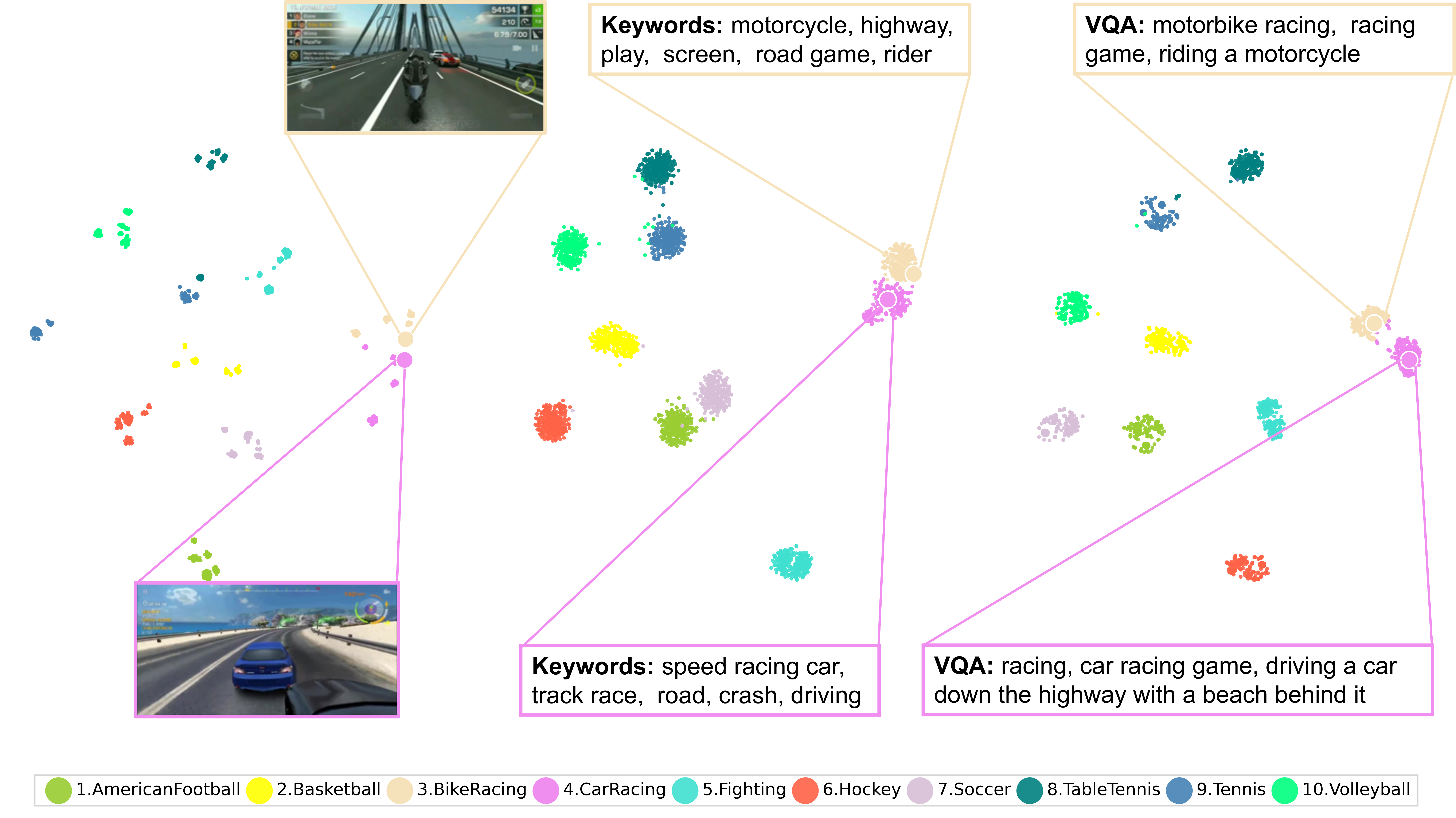}
    \caption{t-SNE embeddings of BLIP2 for the Sports10 dataset. From left to right: Image embedding (Acc: 91.31), Keyword SBERT embedding (Acc: 96.89) and VQA SBERT embedding (Acc: 99.00 with prompt: ``What type of sport is shown in the picture?''). The improvement in cluster accuracy corresponds to better separated clusters in the t-SNE embeddings. }\label{fig:image_caption_carracing_bikeracing}
\end{figure*}

In order to get a better understanding of the comparison of embedding structure, and how generated text relates to that, we provide two examples.
In Figure \ref{fig:image_caption_classroom_conferenceroom} there is an example of the LSUN dataset and in Figure \ref{fig:image_caption_carracing_bikeracing} there is a corresponding example of the Sports10 dataset.

\begin{table*}%[h]
    \centering
    \resizebox{.99\textwidth}{!}{

\begin{tabular}{ll | cc |cc | cc}
\toprule
        &                                     & \multicolumn{2}{c}{Image} & \multicolumn{2}{c}{TF-IDF} & \multicolumn{2}{c}{SBERT} \\
Dataset & Modality / Question & Acc &    NMI & Acc &    NMI & Acc &    NMI \\
\midrule
Sports10 & Image &    91.31 &  93.22 &          &        &          &        \\
        & Caption &          &        &    99.38 &  98.65 &    99.07 &  98.47 \\
        & Keyword &          &        &    99.08 &  97.82 &    96.89 &  96.87 \\
        & Which sport is shown in the picture? &          &        &    84.89 &  94.57 &     98.7 &  98.12 \\
        & What type of sport is shown in the picture? &          &        &    84.83 &  94.46 &     99.0 &  98.21 \\
        & Which game is shown in the picture? &          &        &     84.0 &  90.64 &    95.77 &  95.58 \\
        & Which sports contest is shown in the picture? &          &        &    84.76 &  93.06 &    98.64 &   97.7 \\
\midrule
iNaturalist2021 & Image &    44.97 &   62.7 &          &        &          &        \\
        & Caption &          &        &    34.17 &  39.07 &    47.43 &  61.63 \\
        & Keyword &          &        &    42.13 &  48.25 &    48.44 &  59.48 \\
        & What type of biological object is shown in the picture? &          &        &    38.01 &  47.61 &    47.14 &  61.21 \\
        & What is the biological classification of the object in the picture? &          &        &    35.23 &  39.66 &    47.82 &  60.43 \\
        & Which biological category is shown in the picture? &          &        &     42.1 &   50.3 &    48.57 &  62.23 \\
        & Which species is shown in the picture? &          &        &    45.57 &  38.13 &    45.65 &  56.55 \\
\midrule
LSUN & Image &    62.07 &  64.47 &          &        &          &        \\
        & Caption &          &        &    76.69 &  71.05 &    81.11 &  74.37 \\
        & Keyword &          &        &     76.2 &  69.28 &    70.63 &  70.82 \\
        & What location is shown in the picture? &          &        &    47.04 &  45.12 &    53.49 &  49.11 \\
        & What kind of environment is shown in the picture? &          &        &    72.63 &  67.52 &    81.37 &   74.6 \\
        & What room is shown in the picture? &          &        &     66.4 &  59.92 &    71.59 &  63.54 \\
        & What scene is shown in the picture? &          &        &    76.71 &   70.5 &    78.15 &  77.05 \\
\midrule
HAR & Image &    52.65 &  47.06 &          &        &          &        \\
        & Caption &          &        &    50.51 &  46.09 &    50.85 &  46.68 \\
        & Keyword &          &        &    51.35 &  45.47 &    55.66 &  50.07 \\
        & What type of motion is depicted in the picture? &          &        &    42.68 &  36.69 &     49.2 &  42.54 \\
        & Which activity is shown in the picture? &          &        &    50.77 &  46.04 &    56.03 &  49.69 \\
        & Which action is shown in the picture? &          &        &    52.75 &  48.13 &    58.68 &  52.86 \\
        & What is the person doing in the picture? &          &        &    52.74 &  47.96 &    60.93 &  52.94 \\
\midrule
FER2013 & Image &    35.97 &   21.2 &          &        &          &        \\
        & Caption &          &        &    31.86 &   6.89 &    38.21 &  20.53 \\
        & Keyword &          &        &    47.05 &  27.34 &    46.44 &  29.96 \\
        & What type of countenance is shown in the picture? &          &        &    30.53 &   9.64 &    33.53 &  17.34 \\
        & Which emotion is shown in the picture? &          &        &    46.86 &  34.25 &     45.6 &  36.04 \\
        & Which facial expression is shown in the picture? &          &        &    48.93 &  33.55 &    52.85 &   39.0 \\
        & Which mood is shown in the picture? &          &        &    46.89 &  28.66 &    45.54 &  31.03 \\
\bottomrule
\end{tabular}
    }
    \caption{Full evaluation table for all prompts. All representations, image and text are based on the BLIP-2 model.}
    \label{tab:full_questions}
\end{table*}

\newpage
\section{Explainability}
\label{appendix_sec:explainability}

In this section, we provide pseudo-code for the algorithm in section \ref{subsec:explainability}. As described previously, it counts the number of keyword occurrences per cluster. Afterwards, it takes the top two exclusive keywords.

\begin{algorithm}
\scriptsize
\caption{Explainability}\label{alg:xai_algorithm}
\begin{algorithmic}[1]
\Require \\
$X = \{X_1, X_2, ..., X_m \}:$ be the set of keyword lists for each sample, \\
$Y = \{Y_1, Y_2, ..., Y_m \}:$ be the set of (predicted) cluster labels for each sample, \\
$n:$ Number of output keywords per cluster.
\Ensure List
\Procedure{SimpleXai}{$X, Y$}
\State A, O $\gets$ [], [] \Comment{Active keywords, and others}
\For{\texttt{$i$ in unique($Y$)}} 
\State $K \gets$ count-ordered list of keywords cluster $i$
% \State $A[i]$ $\gets$ top $n$ keywords 
% \State $O[i]$ $\gets$ rest
\State $A[i]$ $\gets$  $K[0:n]$
\State $O[i]$ $\gets$ $K[n:]$
\EndFor
\While{$\bigcap_{i} A [i] \neq \emptyset$}  \Comment{Remove duplicates}
\State $D \gets \bigcap_{i} A[i]$
\State $A[i] \gets A[i] \setminus D$ 
\State $A[i] \gets A[i] \cup O [0: |D|]$
\State $O[i] \gets O[2|D|:]$
\EndWhile
\State \textbf{return} $A$
\EndProcedure%
\end{algorithmic}
\end{algorithm}

\end{document}